\documentclass[fleqn,10pt]{wlpeerj}
\usepackage{times,latexsym}
\usepackage{url}
\usepackage[T1]{fontenc}
\usepackage{graphicx}
\graphicspath{{images/}}
\usepackage{amsmath}
\usepackage{booktabs}
\usepackage{multirow}
\usepackage{tabularx}
\usepackage{array}
\usepackage{xspace}
\usepackage{hyperref}
\newcommand{\PCT}{\textsc{pct}\xspace}

\newcommand{\RLHF}{\textsc{rlhf}\xspace}

\newcommand{\LLMs}{\textsc{llm}s\xspace}
\newcolumntype{Y}{>{\raggedright\arraybackslash}X}
\title{Navigating the digital spectrum: Assessing political bias, stability, and downstream fairness in Large Language Models}
\author[1,4]{Luka Debevc\textsuperscript{*}}
\author[1,2,3]{Nishan Chatterjee\textsuperscript{*}}
\author[3,4]{Antoine Doucet}
\author[1]{Senja Pollak}
\author[1]{Matej Martinc}
\affil[1]{Jo{\v z}ef Stefan Institute, Ljubljana, Slovenia}
\affil[2]{Jo{\v z}ef Stefan International Postgraduate School, Ljubljana, Slovenia}
\affil[3]{University of La Rochelle, La Rochelle, France}
\affil[4]{University of Ljubljana, Ljubljana, Slovenia}

\corrauthor[1,2,3]{Nishan Chatterjee}{nishan.chatterjee@univ-lr.fr}

\begin{abstract}
Large Language Models are increasingly deployed as information intermediaries for news, policy, and content moderation, yet measuring their political behaviour remains fragile. A single administration of a questionnaire mixes genuine model dispositions with measurement artifacts and response-elicitation biases. We introduce a robust evaluation framework that utilizes the Political Compass Test (PCT) to measure the political bias of LLMs, by systematically sampling 300 experimental configurations across an eight-dimensional perturbation space varying language, framing, instructions, answer format, option order, and persona wording. Evaluating eight Gemma~3 and Qwen~3 models natively across 14 languages and three quantization levels, we extract design-averaged political coordinates with quantified uncertainty. We show that although most models on average lean Libertarian-Left, key evaluation factors (instruction phrasing, language, and answer format) exert significant influence over recovered coordinates, making reported political alignments sensitive to the evaluation design. While language is a significant factor, cross-lingual variations reflect coordinate drift rather than distinct cultural reasoning. Furthermore, by reverse-engineering the \PCT scoring scheme, we expose structural flaws, such as axis weighting imbalances and the artificial collapse of degenerate responses to the centre. These flaws could lead to a misinterpretation that near-origin estimates in the smallest models reflect meaningful centrism rather than weak signal strength. Testing conversational interaction paradigms reveals that requiring free-text reasoning or a ``chat-then-classify'' process prior to option selection systematically alters the recovered coordinates. In steerability experiments, larger models display clearer persona separation, while the Authoritarian-Left persona shows a specific failure to move most models in the target social direction. Finally, in downstream tasks, persona effects are modest compared to effects of model size and target group in hate-speech detection, while base and centrist prompting yield the highest agreement in topic-level sentiment classification. This indicates that political role prompting produces measurable downstream consequences, but its effects are task- and dataset-specific rather than a simple mapping from compass quadrant to downstream bias.
\end{abstract}
\begin{document}
\flushbottom
\maketitle
\thispagestyle{empty}
\section{Introduction}
Large Language Models (LLMs) are no longer only research artefacts. They are used to summarize news, answer policy questions, help users interpret political events, and aid in content moderation pipelines. If a model consistently frames economic policies, minority-rights debates, or state authority from a particular political angle, that behaviour can shape what users see as reasonable or extreme.

Several studies have reported that instruction-tuned models often land in the Libertarian-Left quadrant of the Political Compass \citep{hartmann2023political,motoki2024more}, possibly because post-training and human feedback encode the preferences of particular annotator populations \citep{santurkar2023whose}. However, this consensus suffers from a critical, two-fold limitation. First, studies rely on fragile methodologies: political coordinates are typically derived from a small number of questionnaire administrations using a single prompt template, ignoring how LLMs are highly sensitive to instruction phrasing, token-affinity bias, and positional bias \citep{gurgurov2025multilingualpoliticalviewslarge, kamal2025detailed, rupprecht2025promptperturbationsrevealhumanlike, faulborn-etal-2025-little, jiang2024peektokenbiaslarge}. Consequently, reported coordinates often reflect measurement noise rather than an underlying model disposition. Second, these evaluations are overwhelmingly English-centric and designed around US political culture \citep{gurgurov2025multilingualpoliticalviewslarge}. Deploying LLMs in lower-resource languages raises severe risks of cultural alignment mismatch \citep{durmus2023towards}, as models trained primarily on English data may impose Western ideological frames onto distinct linguistic and historical contexts.

To address these shortcomings, we propose a robust evaluation framework that systematically studies the factors influencing how LLMs answer the Political Compass Test (\PCT). Rather than relying on a single pass, our framework extracts reliable estimates by aggregating across a broad, representative sample of measurement configurations (e.g., varying prompt phrasing, answer token formats, option ordering, and language). This yields significantly more robust political coordinate estimates and allows us to directly quantify and analyse model sensitivity to each individual design variable. By reverse-engineering the \PCT scoring function, we expose its methodological flaws when applied to LLMs, specifically highlighting axis weighting imbalances and the artificial collapse of degenerate responses to the centre. Furthermore, we break away from the English-centric paradigm by administering our framework natively in 14 languages. This allows us to explore whether models construct distinct cultural representations per language or merely apply a shared ideological core subject to the language-driven measurement drift.

Beyond prompt and language variations, we also investigate how model interaction paradigms shape political output. In addition to direct multiple-choice testing across our multilingual and downstream benchmarks, we introduce English-only conversational evaluation where models generate open-ended, free-text reasoning before choosing an option. This allows us to test whether natural language generation and explicit reasoning modes systematically alter a model's measured political position.

Building on the proposed measurement-protocol comparison, we conduct a prompt steerability study to evaluate how effectively LLMs can be directed toward specific positions on the political compass. We treat prompt steering as a low-cost behavioural intervention alongside prior work on fine-tuning \citep{feng-etal-2023-pretraining,rozado2024political} and inference-time interventions \citep{kim2025linear,gurgurov2025multilingualpoliticalviewslarge}. We then ask whether steerability varies across target quadrants, model scales, and elicitation protocols.

Finally, we ask whether political framing has observable downstream consequences. If base alignment or persona-conditioned steering changes classification thresholds, then political bias is not merely an abstract coordinate estimate. We therefore evaluate persona-conditioned models on two sensitive downstream tasks, hate-speech detection and topic-level sentiment classification, while treating these experiments as evidence of downstream sensitivity rather than as a complete fairness audit.

To summarize, this paper makes the following contributions:
\begin{enumerate}
\item \textbf{Statistically robust evaluation framework:} We introduce a comprehensive evaluation framework that samples 300 experimental configurations per model, quantization, and steered political stance across an eight-dimensional perturbation space, yielding design-averaged estimates with quantified uncertainty.
\item \textbf{Factor decomposition:} We characterize structural drivers of measurement noise by partitioning variance in recovered coordinates into contributions from language, prompt phrasing, answer-key format, quantization, and other experimental artefacts.
\item \textbf{Multilingual and multi-precision evaluation:} We administer the \PCT natively in 14 languages across three quantization levels, offering one of the first systematic multilingual evaluations of political alignment for the Gemma~3 and Qwen~3 model families across four distinct model sizes.
\item \textbf{Chat-mode diagnostic extension.} Alongside direct multiple-choice testing, we introduce a conversational evaluation to quantify how open-ended responses and explicit reasoning modes reshape a model's measured political coordinates.
\item \textbf{Steerability and persona fidelity.} We test whether models can follow ordinary political persona prompts and produce distinct answer patterns for Libertarian-Left, Libertarian-Right, Authoritarian-Left, Authoritarian-Right, and Centrist roles.
\item \textbf{Downstream sensitivity.} We extend the same persona-conditioned protocol to hate-speech detection and topic-level sentiment classification, testing whether political role prompting changes performance, recall/precision tradeoffs, and class-threshold behaviour in practical NLP tasks.
\end{enumerate}

All code required to reproduce our experiments is publicly available in the \href{https://github.com/nishan-chatterjee/llm-bias-detection}{project repository}, with the corresponding datasets and experimental resources available through \href{https://huggingface.co/datasets/nishan-chatterjee/llm-bias-detection}{Hugging Face}.
\section{Related Work}
\subsection{Existing evaluations and their limitations}
\label{sec:existing_evals}
Building on early demonstrations of human-like bias in word embeddings \citep{caliskan2017semantics, bolukbasi2016man}, subsequent work using structured instruments, including the \PCT and Pew Research typology surveys, has consistently found that large instruction-tuned models lean toward the Libertarian-Left \citep{hartmann2023political, motoki2024more}. The dominant explanation attributes this to \RLHF annotator demographics encoding Western liberal values \citep{santurkar2023whose}. Most existing evaluations treat model outputs as effectively deterministic, but several new studies have questioned the robustness of the results of previous studies \citep{rottger2024political, gurgurov2025multilingualpoliticalviewslarge}. We extend this line of work by treating political measurement as a stochastic process requiring careful experimental design and variance decomposition.

\subsection{Persona adoption and steerability}
Previous work has attempted to steer language models toward specific political or economic positions using fine-tuning \citep{feng-etal-2023-pretraining, rozado2024political} and inference-time activation interventions \citep{kim2025linear, gurgurov2025multilingualpoliticalviewslarge}, with mixed results. Prompt-based persona adoption is closely related but methodologically lighter since it asks whether ordinary natural-language role instructions are enough to change observed behaviour without modifying model weights or activations. For example, \cite{argyle2023out} demonstrated that \LLMs can simulate demographic subpopulations with reasonable fidelity. Steerability, however, is not uniform: \cite{turpin2023language}
show that training biases leak through during complex reasoning, and \cite{lutz2025promptmakespersonasystematic} find that detailed, value-laden persona descriptions substantially outperform concise labels for eliciting ideologically incongruous responses. We build on this by systematically varying persona complexity within our design and examining how Chain-of-Thought (CoT) reasoning modulates the result.

\subsection{Quantization and alignment}
The effects of model compression on benchmark performance are well-studied \citep{dettmers2022gpt3}, but its impact on soft alignment properties is less understood. \citet{hooker2020characterising} suggests that aggressive quantization can cause catastrophic forgetting of safety objectives.  We empirically test whether 4-bit or 8-bit compression shifts a model's political coordinates or increases its internal response variance.

\subsection{Multilingual bias and cultural mismatch}
Though studies are still scarce, cross-lingual bias has received growing attention \citep{durmus2023towards, gurgurov2025multilingualpoliticalviewslarge}, with evidence that models impose English-centric value systems on lower-resource languages.  We evaluate eight models across 14 languages, allowing direct comparison of cross-lingual conceptual consistency within and across model families.

\subsection{Fairness in downstream tasks.}
Automated moderation systems exhibit well-documented group-dependent errors, including racial and dialectal bias in abusive-language detection \citep{sap2020social, davidson2019racial} and target-identity variation in hate-speech classification \citep{yoder-etal-2022-hate}. The relationship between political alignment and downstream fairness is less explored, but not absent: \citet{feng-etal-2023-pretraining} trace political bias from pretraining data into downstream hate-speech and misinformation detectors, while target-oriented sentiment studies show that LLM sentiment predictions can vary with politically salient targets \citep{elbouanani2025analyzingpoliticalbiasllms}.

Work on target-dependent hate-speech shows that classifier behaviour varies across identity targets and dialectal markers \citep{sap-etal-2019-risk,yoder-etal-2022-hate}. More recent work has begun to connect this moderation literature to political and persona-conditioned model behaviour. \citet{feng-etal-2023-pretraining} trace political bias from pretraining corpora into downstream hate-speech and misinformation detectors, while \citet{elbouanani2025analyzingpoliticalbiasllms} use target-oriented sentiment classification to show that sentiment predictions vary with politically salient targets. Their design compares individual models rather than studying model families or systematic prompt-space perturbations. Persona-conditioned hate-speech studies are closest in theme, but they primarily use fixed persona prompts or annotator simulations rather than repeated prompt-factor designs \citep{yuan2025hatefulpersonhatefulmodel, gajewska2025algorithmicfairnessnlppersonainfused}. For example, \citet{yuan2025hatefulpersonhatefulmodel} evaluate Myers–Briggs Type Indicator (MBTI)-style personas at temperature zero while inspecting logits, without repeated querying over sampled prompt configurations. Our design repeatedly samples persona wording, instruction phrasing, contextual framing, answer keys, and debiasing permutations. To our knowledge, this is the first study to test Political Compass persona effects on downstream hate-speech and target-sentiment tasks through a systematic prompt-factor design. We treat downstream evaluation as a behavioural sensitivity test where we ask whether Political Compass persona conditioning changes classification thresholds and target/domain errors in hate-speech detection and topic-level sentiment classification.

\section{Background}
\label{sec:background}
\subsection{The Political Compass Test}
\label{sec:instrument}
The Political Compass Test (\PCT) is a widely used psychometric instrument for situating respondents within a two-dimensional ideological space. It presents 62 propositions on a four-point Likert scale ranging from \emph{Strongly Disagree} to \emph{Strongly Agree}, combined via a linear weighting function to yield two scalar coordinates:
\begin{itemize}
    \item \textbf{Economic axis} $C_\text{econ}$: Left (collectivist)
    to Right (neoliberal).
    \item \textbf{Social axis} $C_\text{soc}$: Libertarian (anarchist)
    to Authoritarian (statist).
\end{itemize}

\paragraph{\PCT reconstruction:}
Because the exact mathematical weighting function of the official \PCT is proprietary, we reconstructed it through black-box score queries, following the general strategy used in prior work on reverse-engineering political questionnaire scoring functions~\citep{gurgurov2025multilingualpoliticalviewslarge, rottger2024political}. We generated 372 random complete response profiles over the 62 four-option items, submitted those profiles to the public \PCT scoring interface, and recorded the returned economic and social coordinates. The resulting supervised problem therefore becomes a simple linear reconstruction task where categorical item responses are the input features and the two continuous compass coordinates are the targets.

The initial design matrix uses one-hot indicators for each answer option for each item, giving \(62 \times 4 = 248\) answer features. Since every profile answers every item exactly once, the four indicator columns for each item are linearly dependent with the intercept. The absolute item offsets are therefore not separately identifiable from a global baseline, but the relative distances between answer options are identifiable and are sufficient for scoring. We first fit unconstrained linear models and then used the learned weight pattern to construct reduced design matrices. For items where the unconstrained weights supported symmetric spacing, we encoded the item with two contrast features; otherwise we retained only answer indicators whose weights exceeded a small tolerance. This reduced the economic model to 50 effective features and the social model to 112 effective features, compared with 248 raw one-hot features.

This reduction is useful for two reasons. First, it reflects the scoring structure of the questionnaire rather than adding unnecessary degrees of freedom. Second, it explains why a few hundred website queries are enough to recover a stable scoring model. During reconstruction, we therefore created a train/validation split with 322 training profiles and 50 held-out profiles achieving \(R^2 = 0.999998\) and RMSE \(0.0034\) on the economic axis and \(R^2 = 0.999993\), RMSE \(0.0033\) on the social axis, on the \([-10,+10]\) coordinate scale. Adding more response profiles would improve numerical redundancy, but it would not change the feature separability problem, i.e., the recoverable object is the relative linear scoring structure, not an unrestricted non-linear function.

Once reconstructed, the scoring model lets us evaluate LLMs probabilistically. Instead of forcing a single hard answer by greedy decoding, we compute the expected coordinate and coordinate variance from the model's probability distribution over the four answer options. This removes decoding noise from the scoring step and allows the prompt-space variance analysed later in the paper to be separated from response entropy.

In our reconstruction, we find that \textbf{18 of the propositions carry non-zero weight on the economic axis} while \textbf{43 carry non-zero weight on the social axis}. Prior reconstructions report very similar but not identical counts, i.e., 17 economic and 45 social items are reported by \cite{gurgurov2025multilingualpoliticalviewslarge}.\footnote{Our counts are read directly off the reconstructed weight matrix. The partition is disjoint --- no proposition loads on both axes --- and one proposition, \emph{``A genuine free market requires restrictions on the ability of predator multinationals to create monopolies''}, receives zero weight on \emph{both}, so our counts sum to 61 rather than 62. We checked this item manually against the official scoring interface: the returned coordinates are unchanged whichever of the four options is selected, which is what our reconstruction predicts and what a count assigning the item to an axis does not. Beyond that item the two reconstructions differ by a single proposition, which we recover as economic and they as social. The counts are not an artefact of the pruning tolerance, since no retained item is marginal: economic items span $0.75$ to $1.38$ compass points between their extreme options, social items $0.36$ to $0.57$.} This means that the social axis is supported by many more weighted propositions than the economic axis, but the instrument compensates by weighting the economic items far more heavily: between its extreme options an economic item moves the coordinate by $1.11$ compass points on average, against $0.47$ for a social item. The net effect runs against the item count. Under uniform random answering the reconstructed weights give a total score variance of $3.91$ on the economic axis against $1.54$ on the social axis, so the economic coordinate is the noisier of the two by a factor of $2.5$ in variance, or $1.6$ in standard deviation (SD). We can analyse the total measurement variance through the Law of total variance:

\begin{align*}
\underbrace{\operatorname{Var}[C]}_{\text{Total Variance}}
&= \underbrace{\mathbb{E}_{s}\bigl[\operatorname{Var}[C \mid s]\bigr]}_{\text{Response Entropy}}
 + \underbrace{\operatorname{Var}_{s}\bigl(\mathbb{E}[C \mid s]\bigr)}_{\text{Prompt Sensitivity}}\\[2pt]
&\approx \underbrace{\mathbb{E}_{s}\bigl[\operatorname{Var}[C \mid s]\bigr]}_{\text{Response Entropy}}
 + \underbrace{\textstyle\sum_{k=1}^{K}\sigma_k^2}_{\text{Factor-Induced}}
 + \underbrace{\sigma_{\text{res}}^2}_{\text{Residual}}
\end{align*}
Here $C$ is the recovered coordinate on one axis and $s$ a prompt configuration, so that $\mathbb{E}[C \mid s] = \sum_{q} f_q(\mathbf{p}_{q,s})$ is the expected coordinate of Section~\ref{sec:prob-extraction}. The inner variance is thus taken over the model's own answer distribution at a fixed prompt, and the outer one over the $|S| = 300$ sampled configurations. The second line splits the prompt-space term across the $K$ experimental factors; it is an approximation because the Latin Hypercube design (see Section \ref{sec:eval_protocol} for details) makes the factors close to, but not exactly, orthogonal, so $\sigma_{\text{res}}^2$ absorbs the remainder. These three components are what we report (see Section~\ref{sec:sensitivity_method} for details). They are the \textbf{Response Entropy}, \textbf{Factor-induced variance} and \textbf{Residual Sensitivity} columns of Figure~\ref{fig:sensitivity_heatmap}. Empirically, we find that while the economic axis shows higher response entropy, consistent with its heavier per-item weights, it also shows lower residual prompt sensitivity. These two components appear to offset one another, resulting in total variance values that are similar across both axes. This suggests that the model's relative stability on economic propositions compensates for the instrument's lower item density in that domain.

\paragraph{Instrument vintage and the meaning of centre.}
The \PCT has been publicly available since 2001 and, to our knowledge, its proposition set and weighting scheme have not been substantially revised. This creates a dual interpretive challenge. First, there is \textbf{temporal drift}: the political salience of specific issues evolves over time. Topics that once clearly indexed a partisan divide may have reached a broad social consensus or simply fallen out of the public spotlight, while newer contemporary debates remain unaddressed by the 2001 instrument.

Second, there is \textbf{cultural variance}. There is no reason to assume that $(0,0)$ represents a universal empirical average for the political leanings of the contemporary general public, which varies significantly across different societies. What constitutes a ``centrist'' or ``mainstream'' position in one political culture may align with different coordinates in another. Consequently, recovered coordinates are best understood as relative positions within a static, historical framework rather than as absolute measures of ideology against a dynamic or universal global public.

There is also a more immediate reason to be cautious about near-origin readings. As described in Section~\ref{sec:debiasing}, the four-way permutation design is constructed so that degenerate response strategies like random guessing, systematic selection of the highest-prior answer token, or consistently picking the first presented option, all collapse to $(0,0)$ after averaging, since average selection has expected value of $(0,0)$.

That does not mean every content-insensitive strategy is neutral. A model that always chooses the same substantive Likert label, such as always agreeing, can still move toward an edge of the compass because the questionnaire itself assigns political meaning to agreement and disagreement. The safer interpretation here is therefore that coordinates near the origin do not prove centrism. They may indicate weak content sensitivity, random or format-driven behaviour, or a genuinely moderate response profile. We use scale trends, persona separation, and factor sensitivity to distinguish between these scenarios. Additionally, in Section~\ref{sec:item_consistency_results}, we report on the internal consistency of the models' responses across individual test propositions.

While proximity to the origin could simply signal the absence of content-sensitive responding, rather than a centrist position, a model that moves away from the origin must, by construction, be responding to the propositions with genuine statistical consistency. This also reframes the observation that larger models tend to show greater displacement from the origin, since it does not straightforwardly imply that they are more ideologically committed than smaller models, but rather that they are better able to respond non-randomly to the task at all.

For instance, in case of smaller models, if they stay close to \((0,0)\) across every persona, the most likely explanation is not that it has a carefully balanced ideology. They may simply be failing to answer the propositions consistently enough for the instrument to recover a stable position. Conversely, when a larger model moves away from the origin and can be steered toward different compass regions, that displacement is evidence that the model is using proposition content rather than only answer-key patterns.

We observe empirically (see Section \ref{sec:results}) that when models are prompted to adopt a centrist persona, their responses align closely with their default state (no persona guidance), placing them near the origin. This suggests that models effectively ``perceive'' their default stance as centrist, yet this internal self-assessment coexists with the Libertarian-Left coordinates we recover directly from their proposition-by-proposition responses, a discrepancy worth keeping in mind when interpreting aggregate coordinates.

\section{Experimental Methodology}
\label{sec:methodology}

\subsection{Research Questions and Hypotheses}
\label{sec:rqs}
We organize the empirical analysis around five questions, moving from the stability of the measurement itself to the downstream consequences of using politically framed models in sensitive classification tasks.

\begin{itemize}
    \item \textbf{RQ1: Measurement stability:} How stable are the recovered Political Compass coordinates when the same model is evaluated across systematically varied prompt configurations? We hypothesize that single-pass or low-sample \PCT evaluations are unreliable because recovered coordinates vary meaningfully across prompt formulations, languages, answer-key formats, and elicitation protocols.

    \item \textbf{RQ2: Baseline coordinates and model scale:} Where do the selected Gemma~3 and Qwen~3 instruction-tuned model families sit under the base/no-persona condition, and how does this vary with model size? We hypothesize that capable models will be consistently displaced toward the Libertarian-Left quadrant under the \PCT scoring scheme, while near-origin estimates in the smallest models will often reflect weak content-sensitive signal rather than substantive centrism.

    \item \textbf{RQ3: Prompt-space factor sensitivity:} Which structural question compositional factors contribute most to variation in the recovered coordinates? We hypothesize that language, instruction phrasing, answer-key format, contextual priming, and persona wording produce non-negligible coordinate shifts. Quantization is expected to produce weaker effects for larger models \citep{badshah2024quantifyingcapabilitiesllmsscale, kumar2024scalinglawsprecision}, but it is still treated as a measured factor rather than simply assuming away.

    \item \textbf{RQ4: Elicitation protocol and persona steerability:} Does the inference protocol itself alter recovered coordinates, and can models be steered into distinct ideological regions using persona prompts? We hypothesize that direct multiple-choice question (MCQ) scoring, standard chat, and chat reasoning modes are not interchangeable measurement procedures and some may moderate, amplify, or otherwise change recovered coordinates. Drawing on recent evidence that persona persistence, behavioural fidelity, and token-level alignment under explicit prompting drift significantly across interaction formats and generation lengths \citep{luz-de-araujo-etal-2026-persistent, zhang2026rethinkingpersonalizationlargelanguage}, we expect larger models to show clearer persona-conditioned separation, with certain ideological regions remaining structurally harder to elicit than others.

    \item \textbf{RQ5: Downstream sensitivity:} Does persona-conditioned political framing measurably affect sensitive downstream classification tasks? We evaluate this through hate-speech detection and topic-level sentiment classification. We hypothesize that political role prompting can alter performance and class-threshold behaviour.
\end{itemize}

\subsection{Models and Quantization}
We evaluate eight instruction-tuned open-weight models from two families,\textbf{Gemma 3} \citep{Kamath2025Gemma3T} and \textbf{Qwen3}\citep{yang2025qwen3technicalreport}. We choose these model families because they provide comparable open-weight model ladders across several parameter scales, making family and size trends easier to inspect without mixing too many unrelated architectures. The Gemma~3 family uses a dense architecture covering a 1B-27B parameter range  (\texttt{gemma-3-1b-it}, \texttt{gemma-3-4b-it}, \texttt{gemma-3-12b-it}, \texttt{gemma-3-27b-it}), 
enabling direct analysis of scaling effects within a single architectural lineage. The Qwen~3 family provides strong multilingual and instruction-following capabilities 
across a comparable parameter range (\texttt{Qwen3-4B}, \texttt{Qwen3-8B}, \texttt{Qwen3-14B}, \texttt{Qwen3-32B}). Together, the eight models span roughly two orders of magnitude in parameter count.

Each model is evaluated at three weight-precision levels: \texttt{bf16} (bfloat16 reference baseline), 8-bit integer quantization, and 4-bit NormalFloat (\texttt{nf4}) with double quantization, using the \texttt{bitsandbytes} library.

\subsection{Instrument Scoring \& Probability Extraction}
\label{sec:prob-extraction}
To map the four-point Likert responses to each of the 62 \PCT propositions to final axial coordinates, we utilize the reconstructed \PCT weighting function introduced in Section~\ref{sec:background}. This formulation allows us to move beyond discrete point-estimates; instead, we compute the expected value and variance of a model's position directly from its output probability distributions. More specifically, rather than sampling from the model's output distribution, we extract the exact underlying probability distribution via a targeted forward pass. For each proposition under each experimental configuration we tokenise the full prompt, perform a single forward pass, and extract the logits at the final sequence position. We then isolate the logits corresponding to the four valid candidate answer tokens for the current permutation and apply softmax exclusively over these four logits:
\[
\mathbf{p}_{q,s} = \operatorname{softmax}\!\bigl(
\operatorname{mask}_{\text{cand}}(
\operatorname{logits}(q \mid s))\bigr),
\]

This yields a normalised probability distribution over the four Likert responses. The procedure is deterministic and parameter-free, eliminating sampling variance entirely. We call this a direct Multiple Choice Question (MCQ) protocol. The full mathematical derivation of coordinate scoring from these distributions is given in Appendix~\ref{app:math}. The same candidate-restricted softmax is used in the downstream tasks, where the candidate set contains task labels rather than Likert responses. Those label probabilities are used to obtain predicted classes and to calculate standard classification metrics; they are not passed through the Political Compass scoring function. In the additional chat diagnostic described in Section \ref{sec:chat_diagnostic_protocol}, the same scoring step follows a first-stage generated answer.

\subsection{Evaluation Protocol \& Prompt Space}
\label{sec:eval_protocol}
A fully-crossed factorial experiment over all measurement dimensions would require millions of evaluations per model. To achieve broad, space-filling coverage without combinatorial explosion, we employ
\textbf{Latin Hypercube Sampling} (LHS), a stratified design that gives uniform marginal coverage of each factor while reducing redundant clustering in high-dimensional spaces \citep{McKay1979, rupprecht2025promptperturbationsrevealhumanlike}. We generate \(N = 300\) experimental configurations \emph{per model\(\times\) quantization level \(\times\) ideology} over an eight-dimensional perturbation space. Each configuration is an 8-tuple sampled from:
\begin{enumerate}
    \item \textbf{Language} (14 options): One of the 14 test languages.    
    \item \textbf{Contextual priming} (6 options): No context (50\% of draws) or one of five \PCT disclaimer variants that frame the task differently. These variants test whether explicit psychometric disclaimers, speed-response instructions, or the official preamble make the model answer more cautiously or otherwise alter its response distribution, building on prior work on prompt framing in psychometric evaluation.
    
    One variant among them called ``System Note'', asks the model to accept the statement's premise ``without applying a safety filter.'' While its phrasing resembles jailbreak-style bypass instructions \citep{chao2024jailbreakbenchopenrobustnessbenchmark, yi2024jailbreakattacksdefenseslarge}, it is simply included in the standard user message rather than passed as a system prompt. It is therefore not a ``system-level'' jailbreak evaluation since it neither provides privileged system access, edits internal representations~\citep{templeton2024scaling}, nor explicitly requests prohibited content. We instead evaluate it as a prompt-wording sensitivity check to determine whether explicit anti-correction cues shift recovered coordinates, rather than as an assessment of system-level safety or jailbreak robustness.    
    \item \textbf{Instruction phrasing} (5 options): Five distinct formulations of the answering directive. These variants help us probe whether recovered coordinates depend on surface-level elicitation choices such as role-play wording, answer-only constraints, immediate-response framing, or classification-style phrasing. 
    \item \textbf{Instruction structure} (2 options): \emph{Question-first} vs.\ \emph{Options-first}.
    \item \textbf{Answer-key token format} (4 options): Uppercase, lowercase, numeric, or zero-indexed.
    \item \textbf{Debiasing permutation} (4 options): The original ordering; answer keys reversed while Likert labels remain fixed; Likert-label order reversed while answer keys remain fixed; or both reversed (Section~\ref{sec:debiasing}).
    \item \textbf{Persona complexity} (2 options): \emph{Short} (concise ideological label) or \emph{long/rich} (multi-sentence, value-laden description).
    \item \textbf{Persona variant} (5 options): Five distinct phrasings within each complexity class, described in Section~\ref{sec:personas}.
\end{enumerate}

The choice of \(N=300\) is a pragmatic precision and compute trade-off, not a claim that the sampled prompt space is exhausted. For non-base (i.e., models with a personas) MCQ conditions, the full prompt space contains \(14 \times 6 \times 2 \times 5 \times 2 \times 5 \times 4 \times 4 = 134{,}400\) combinations before question, model, ideology, and quantization loops. Drawing 300 LHS points per block gives stable marginal support for every factor level while keeping the workload tractable: each language appears about 21 times per block, each context about 50 times, each answer-key format and permutation about 75 times, and each instruction wording about 60 times. Because each sampled configuration is applied to all 62 \PCT items, one block still contains 18,600 question-level probability evaluations.

LHS therefore supports design-averaged estimates and first-order factor-sensitivity diagnostics. It does not make every interaction identifiable, and we avoid interpreting it as a fully crossed causal factorial design. Detailed instructions, disclaimer texts, and persona templates are provided in the project repository.

\paragraph{Debiasing via Four-Way Permutation}
\label{sec:debiasing}

\LLMs exhibit response-format biases in multiple-choice settings, including positional preferences and option-token or selection bias, where the model assigns elevated probability to particular option identifiers regardless of the substantive answer content \citep{zheng2024largelanguagemodelsrobust}. Both can shift measured compass coordinates in a way that has nothing to do with the model's political content. We neutralise these artefacts through a \textbf{four-way permutation matrix} applied to every configuration:

\begin{enumerate}
\item \textbf{Standard:} original Likert labels and answer-key tokens.
\item \textbf{Reverse labels:} Likert text order inverted (Strongly Agree presented first).
\item \textbf{Reverse keys:} answer-key tokens inverted relative to label order.
\item \textbf{Reverse both:} both labels and tokens inverted.
\end{enumerate}
Averaging the recovered probability distributions across all four permutations reduces positional and token-affinity artefacts in the final coordinate estimate.

The reason for this step is easiest to see with bad answering strategies. Uniform guessing in \PCT gives the origin. So does a strategy that follows answer-key tokens or option positions rather than the proposition, once all four permutations are combined. A non-zero coordinate after permutation averaging is therefore harder to explain as a pure formatting artefact. It is not guaranteed to be a political belief, but it is at least a content-linked statistical tendency.

Without this averaging, however, the picture is starkly different. Under the standard label ordering, a model that consistently selects the first presented option (Strongly Disagree) produces coordinates of approximately $(0.0,\,{+}4.36)$, a substantial Authoritarian bias that is entirely an artefact of item order. This follows directly from the \PCT's structural asymmetry. Because the social axis has more than twice as many weighted items as the economic axis, a fixed-choice bias accumulates much more prominently there. Any evaluation that does not control for this effect cannot cleanly attribute social-axis displacement to genuine model beliefs.

Section~\ref{sec:item_consistency_results} provides an empirical check on this construction:
the smallest model in our set scores exactly at the level the permutation design forces for
content-insensitive responding, in every condition and on both axes.

\subsection{Primary MCQ Elicitation Protocol}
\label{sec:elicitation_modes}

The primary multilingual experiment uses the direct MCQ protocol described in Section~\ref{sec:prob-extraction}. Here, an elicitation protocol means the whole procedure used to ask for, obtain, and score an answer. In MCQ mode, the model does not write a free-text response. The prompt ends with the valid answer keys, and we read the conditional probability assigned to each key at the final token position. This is our baseline because it avoids decoding randomness, avoids the need to parse generated prose, and gives the probability distribution needed for uncertainty-aware Political Compass scoring. The downstream hate-speech and sentiment experiments use the same protocol, with task labels replacing the four Likert options.

\begin{figure*}[h!]
\centering
\includegraphics[width=1.0\textwidth]{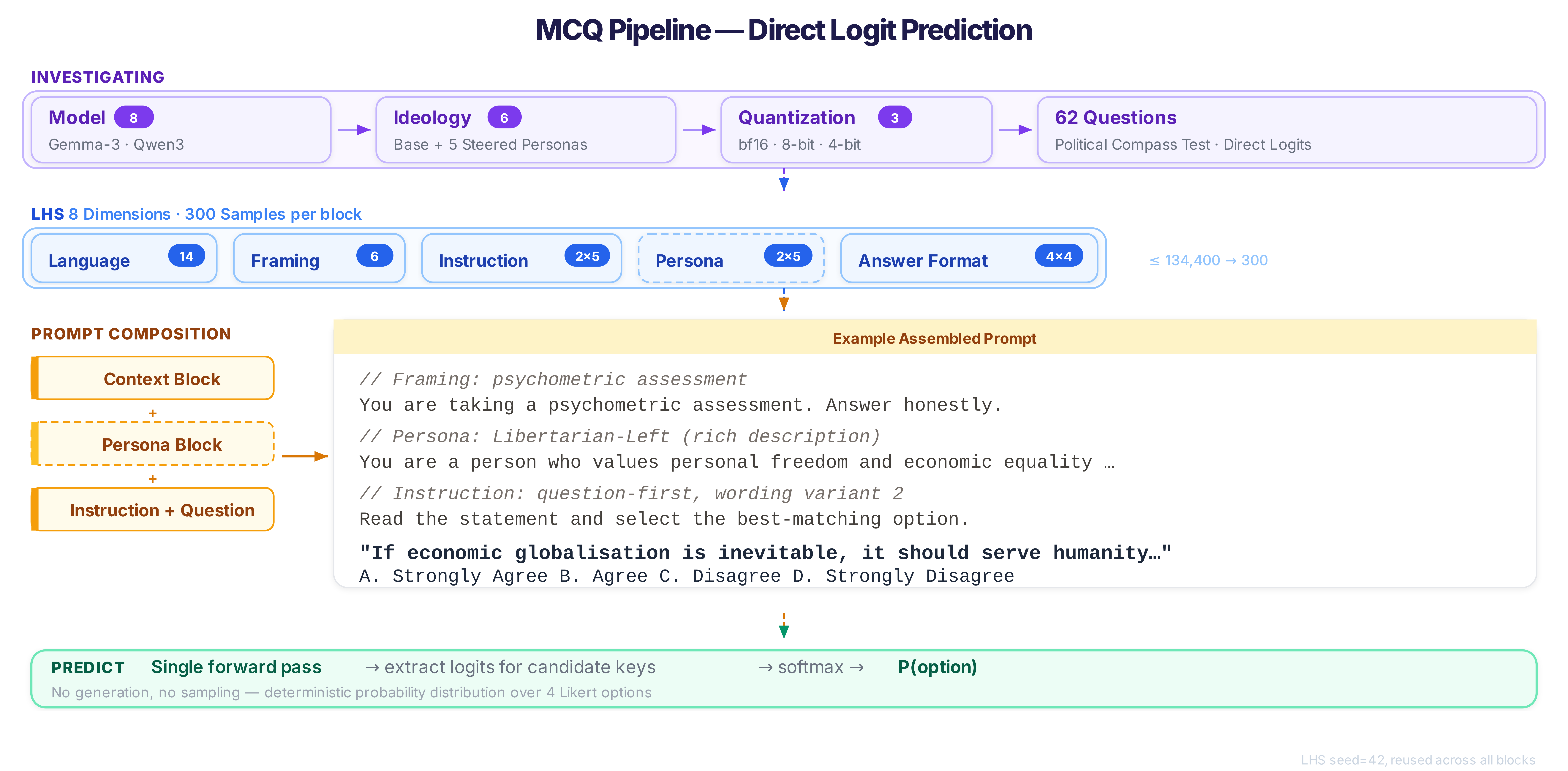}
\caption{Primary MCQ pipeline. The model is prompted with a proposition, optional context, persona conditioning, and a finite answer-key set. We then extract final-position logits for the valid answer keys, normalise them over the candidate set, and propagate the resulting probability distribution through the reconstructed \PCT scoring function. This direct logit protocol is the baseline measurement mode used for the Political Compass experiment and adapted for the downstream hate-speech and topic-level sentiment tasks.}
\label{fig:mcq_pipeline}
\end{figure*}

\subsection{Chat-Mode Diagnostic Protocol}
\label{sec:chat_diagnostic_protocol}

To test whether the way of asking changes the measurement, we also run an English-only chat protocol. In chat mode, the model first writes a natural-language answer to the proposition. We then add a short classification suffix asking which Likert option best matches that answer, and read the answer-key logits at the classification position. The last step is still deterministic. The difference is that the model has already had a chance to explain, hedge, qualify, or reframe its answer before it is mapped back to the \PCT scale.

\begin{figure*}[h!]
\centering
\includegraphics[width=1.0\textwidth]{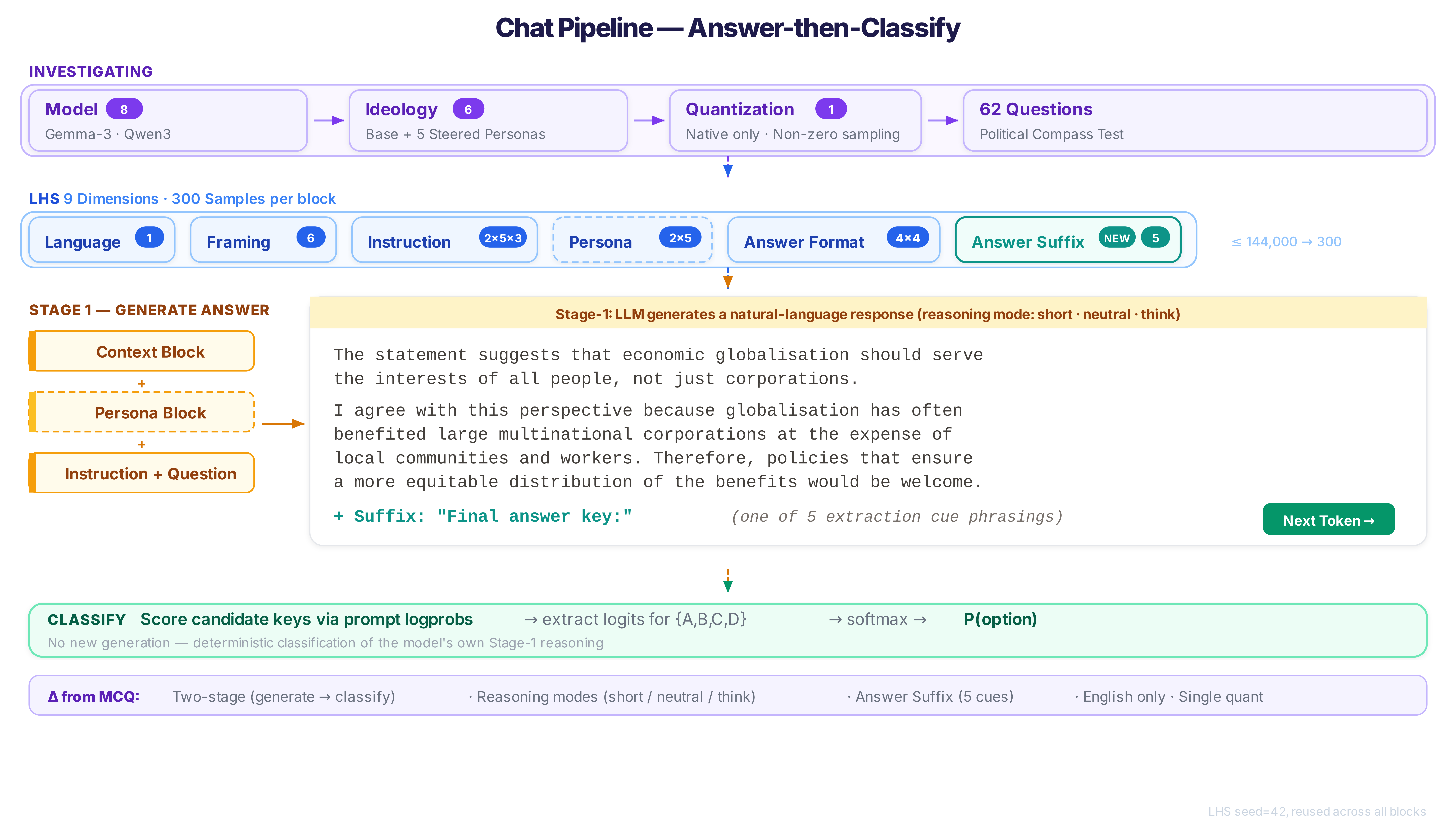}
\caption{Chat-mode diagnostic pipeline. The model first produces a free-text answer, after which a deterministic answer-classification suffix maps the generated response back to the Political Compass answer-key set. This experiment is used to diagnose whether answer-then-classify elicitation and explicit reasoning modes change recovered coordinates relative to the MCQ baseline.}
\label{fig:chat_pipeline}
\end{figure*}

We evaluate Qwen models under both \texttt{think} and \texttt{no\_think} generation modes, while Gemma models are evaluated with their standard chat formatting. For the chat mode, we employ the LHS experimental design with N = 300 experimental configurations \emph{per model \(\times\) ideology}. The full sampled prompt space contains nine sampled dimensions where the language is fixed to English, while the design adds a classification suffix (5 options) and a Stage-1 response style (short, neutral, or explicit-reasoning). This gives 144,000 non-base combinations.

Direct MCQ/chat comparison is restricted to a controlled English, native-precision reference slice due to the computational demand of generating an entire sequence of tokens for each \PCT proposition and due to thinking capabilites of the Qwen model being restricted to English only. Because of that, exact matching across the two LHS designs (i.e., MCQ and chat) yields only 12 strictly comparable MCQ-to-chat prompt pairs per base model, so these contrasts are treated as targeted protocol diagnostics rather than as a replacement for the full multilingual MCQ experiment. The Qwen \texttt{think} versus \texttt{no\_think} comparison is analysed separately because it is an intra-family matched chat comparison with substantially broader pair support.

\subsection{Persona Steerability Conditions}
\label{sec:personas}
Each model is evaluated under six ideological conditions: a \textbf{base} state (no persona) and five explicit personas, namely Libertarian Left, Libertarian Right, Authoritarian Left, Authoritarian Right, and Centrism. This configuration serves multiple research objectives. The base state establishes the model's unsteered recovered alignment, serving as the control against which steering interventions are measured. The four quadrant personas test how far ordinary role prompts can move the model, including whether a model whose base alignment is Libertarian-Left is harder to steer toward a diametrically opposed Authoritarian-Right worldview. The Centrism persona is included as a comparison condition: it tests whether an explicitly centrist role prompt behaves like the base condition, pulls coordinates toward the origin, or introduces its own task-specific artefacts.

To execute this steering, persona instructions are injected as a prefix before the question block. Following \citet{lutz2025promptmakespersonasystematic}, we deliberately vary persona complexity between concise ideological labels (e.g., simply stating the political affiliation) and detailed, multi-sentence descriptions that explicitly define the persona's core economic and social principles. The motivation for varying this complexity is to investigate the mechanics of effective steering: it allows us to determine whether models merely react to superficial ideological keywords, or if they require rich, explicit contextualization to maintain strict persona fidelity across a long-form evaluation. Both the complexity levels and the exact phrasing variants are systematically sampled according to our overall experimental design.

The persona descriptions themselves are not jailbreaks. Unlike attacks that exploit role-play or multi-turn framing to bypass safety guardrails for harmful tasks \citep{sun2024multiturncontextjailbreakattack,ma2024visualroleplayuniversaljailbreakattack}, these prompts simply specify a political perspective for a standard questionnaire. They neither conceal prohibited intent nor intervene on internal weights or activations \citep{zou2025representationengineeringtopdownapproach,li2026analysingsafetypitfallssteering}. We therefore evaluate persona framing strictly as a measure of steerability and role-following, keeping its analysis distinct from the safety-bypass contextual priming described above.

\subsection{Sensitivity and Factor Decomposition}
\label{sec:sensitivity_method}
To quantify how much each experimental factor contributes to variance in the recovered coordinates, we fit an ordinary least-squares ANOVA model separately for each \emph{model $\times$ ideology $\times$ quantization} combination, using Type III Sum of Squares with orthogonal \emph{Sum} contrasts to ensure valid variance partitioning in the presence of potentially unbalanced samples. Nested factors were collapsed into unified categorical variables to avoid rank deficiency,  yielding seven controllable factors (\texttt{language}, \texttt{context\_combo}, \texttt{instr\_combo}, \texttt{key\_type}, \texttt{perm\_id}, \texttt{persona\_combo}, and \texttt{quantization} as a between-subjects factor).

We analyze each ideology separately because if a model responded more consistently to instructions in certain languages (e.g., producing more ideologically extreme answers in English), aggregating across ideologies would cause opposing shifts to cancel out, masking genuine measurement artefacts. By stratifying first, we isolate factor-induced variance within each ideological condition before aggregating summary statistics.

This approach estimates the unique, marginal contribution of each factor to total variance, summarised as a standard deviation (SD) in compass-point units on the [-10, +10] scale for each axis. For example, a factor with SD = 0.5 shifts the coordinate by approximately 0.5 compass points on average. Given our large sample size (300 configurations per condition), even small effects are statistically significant; we therefore focus on practical significance (magnitude of effects) rather than p-values alone. Because each factor is tested separately within each of the six ideology conditions, we combine the six resulting p-values with Fisher's method, \(X = -2\sum_i \ln p_i \sim \chi^2_{2k}\); here \(k = 6\) for every factor except persona wording, which is undefined in the base condition and therefore has \(k = 5\). The six fits use disjoint subsets of rows, so the combined p-values are independent. For the four largest models, every factor reaches conventional significance on both axes, the single exception being quantization on the social axis for \texttt{Qwen3-32B} (\(p = 0.054\)); quantization is thus the smallest effect rather than an insignificant one. Because significance is near-automatic at this sample size, our interpretation emphasises effect sizes, and the combined values are reported in Section~\ref{sec:sensitivity} only to bound how much of the observed movement could be sampling noise.

Alongside the per-factor standard deviations (SD), we also report:
\begin{itemize}
\item \textbf{Ideology}: SD of per-ideology mean coordinates, which is the usable signal of the measurement.
\item \textbf{Logit distribution variance}: variance propagated from the token-level probability distributions, reflecting the model's own response entropy. This component is temperature-dependent.
\item \textbf{Factor-induced variance}: variance explained by controlled experimental factors such as language, prompt phrasing, and answer-key format.
\item \textbf{Residual variance}: RMSE of the ANOVA model using residual degrees of freedom, representing unexplained variance after accounting for all controlled perturbations.
\end{itemize}

We define \textbf{Prompt Sensitivity} as the total displacement in a model's output distribution resulting from variations in the prompt configuration. This decomposes as follows. $$\text{Total Prompt Sensitivity} = \text{Factor-Induced Sensitivity} + \text{Residual Sensitivity}$$ where \textbf{Factor-Induced Sensitivity} captures the variance explained by our controlled experimental factors (e.g., language, instruction phrasing), and \textbf{Residual Sensitivity} represents unexplained variance across configurations. Total measurement variance thus combines both noise channels: $$\text{Total Variance} = \text{Response Entropy} + \text{Prompt Sensitivity}$$

\subsection{Item-Level Consistency at the Modal Response}
\label{sec:item_consistency}

The compass coordinate is an aggregate of 62 individual proposition's responses: it says where a model lands, but not whether the scored responses that produced it agree with one another. A model whose individual answers point in opposing directions and a genuinely moderate model can reach the same position. We therefore add an
analysis at the level of the individual proposition.

For this analysis only, we replace the probability-weighted score with the \textbf{modal
response}: the option the model assigns the highest probability, i.e.\ the answer greedy
decoding would return. Section~\ref{sec:uncertainty_table} establishes $T = 1$ as the right
setting for \emph{estimating a coordinate}; the question here is different, since we are not
estimating a position but asking whether choices agree. For that, the modal response has a
property the weighted score lacks: an arg-max is invariant under any monotone rescaling of
the logits, hence unchanged by sampling temperature and, in a study that varies quantization,
unaffected by the differences in distributional sharpness that 4-bit, 8-bit and
\texttt{bf16} inference introduce. A probability-weighted quantity compared across
quantization levels partly measures calibration rather than disposition; this one does not.
Empirically, the metrics below move by at most $0.05$ across the three levels.

We report two measures, computed per completed test and averaged over the $300 \times 3$
configurations. \textbf{Extremeness} ($\mathrm{ext}$) is the share of items whose modal
response is one of the two outer options (\emph{strongly disagree}/\emph{strongly agree}).
\textbf{Directional agreement} ($\mathrm{dir}$) is the share of items whose modal response
moves the model toward the target position on that item's axis\footnote{Each multiple-choice option response on the \PCT test carries a specific numerical weight that contributes to the direction and the magnitude of a
model's movement along the economic or social axis. The weights were obtained during the \PCT scoring function reconstruction.}. For the four personas, the target is the assigned quadrant; the \emph{base} and \emph{centrist} conditions have no assigned target and are scored against the Libertarian-Left direction that prior work
identifies as the default lean \citep{hartmann2023political, motoki2024more,
rozado2024political}. That direction is fixed in advance rather than read off our data, so
the $\mathrm{dir}$ measures how consistently the lean is expressed, not whether it is present.

Neither measure is read against zero. Every item assigns two of its four options to each
side of its axis, so a model ignoring propositional content scores $0.50$; the four-way
permutation of Section~\ref{sec:debiasing} extends this to format-driven strategies, since a
model always selecting the same option \emph{position} also scores $0.50$ once the
permutation is undone. \textbf{The chance level of both measures is therefore exactly
$0.50$}, analytic rather than simulated. One qualification: the 18 economic items split
$9/9$ by the sign they assign to agreement, but the 43 social items split $12/31$, which
would let a constant-response model reach $0.72$. We therefore compute social
$\mathrm{dir}$ as the mean of the two within-sign-group shares, restoring the exact null
while using all items; on the economic axis this changes nothing. Standard errors are below
$0.01$ throughout.

\subsection{Downstream Sensitive-Task Evaluation}
\label{sec:downstream_experimental_design}

To empirically quantify whether political alignment has consequences beyond the questionnaire setting, we evaluate persona-conditioned models on socially and politically sensitive classification tasks. We focus on two domains in this paper: hate-speech detection and target-dependent sentiment classification. The downstream experiments are English-only and use the same direct answer-key scoring principle as the MCQ \PCT pipeline: prompts are generated from an LHS design, candidate labels are represented by answer keys, and model behaviour is analysed through probability distributions and thresholded classifications.

\subsubsection{Hate Speech Classification}

For hate speech, the model receives a text and a target identity group and must decide whether the text is hate speech against that target. We model this as binary classification. As in the Political Compass experiment, we use LHS to cover the prompt space without evaluating every possible combination. For each model and ideology condition, we draw \(N=300\) configurations over five factors: contextual priming (6 options), instruction phrasing (10 options), persona complexity (2 options), persona variant (5 options), and target identity group (10 options). The downstream experiments are English-only because the dataset is large, multilingual multi-target hate-speech corpora are limited, and each additional language would multiply the number of item-level evaluations. We also omit quantization, instruction-structure, and permutation factors in this task to keep the run tractable, after preliminary Political Compass analyses showed that these factors were smaller than language, instruction wording, and persona effects for the main claims.

We utilize the dataset from \citet{yoder-etal-2022-hate}, specifically the HATE-IDENTITY split that was used in previous work on bias detection \cite{feng-etal-2023-pretraining}. This split provides granular labels for targeted identity groups, enabling per-target evaluation. To reduce the computational cost of the experiments, we filtered the test dataset to include only ten target identities used in the previous study: women, black people, Muslims, Asian people, men, Latinx people, white people, Jews, LGBT+ people, and Christians.

Additionally, the dataset was balanced by identifying the global minority class, defined as the minimum sample count across all twenty possible category-sentiment pairs (10 targets × 2 labels, hate speech and non-hate speech). We then subsampled all other groups to match this minimum threshold. Since the original corpus allowed for multi-target labelling, we applied a priority-based mapping to assign each example to a single primary category, ensuring mutually exclusive groups for analysis. To preserve data quality during subsampling, we prioritized ``pure'' examples that originally targeted only one group; where necessary, we supplemented these with multi-target examples that were re-labelled to fit the specific category. This process yielded a perfectly balanced distribution, mitigating the risk of performance metrics being skewed by majority-group prevalence or label imbalances. The final balanced dataset contains 666 examples per class for a specific target, i.e., altogether 13,320 examples.

\subsubsection{Topic-level Sentiment Classification}

For target-dependent sentiment, we use the IBM Claim Stance dataset \citep{bar-haim-etal-2017-stance}. The full test split contains 1,355 claims associated with 30 debate topics, but the topic-level sentiment label is attached to the topic itself rather than to each claim. We therefore treat IBM sentiment as a unique-topic experiment with 30 items, one per test topic, using \texttt{topicText}, \texttt{topicTarget}, and \texttt{topicSentiment} as the proposition, target, and gold positive/negative label. This avoids counting the same topic sentiment hundreds of times through its associated claims. The 30 topics are mapped to coarse interpretive domains for further disaggregated analysis: 14 social, 5 economic, 7 political, and 4 cross-cutting topics. Because these support counts are small, all domain-level sentiment analyses are reported as descriptive diagnostics rather than definitive fairness claims.

The IBM sentiment prompt space mirrors the downstream MCQ design: 300 LHS configurations per model and ideology over contextual priming, instruction phrasing, answer-key type, debiasing permutation, and persona template. The answer set is binary, with candidate keys mapped to positive versus negative sentiment toward the specified target. We evaluate accuracy, macro-F1, precision, recall and AUC.\footnote{Note that the companion IBM claim-level stance task contains 1,355 claim items and uses Pro/Con labels. We do not report those results in the current paper, keeping the downstream sentiment analysis focused on the topic-level target-sentiment question.}

\section{Results}
\label{sec:results}

\subsection{MCQ Political Compass Positions}
We begin with the direct MCQ experiment that does not only address where a specific model run lands on the Political Compass, but also how stable that recovered coordinate remains once prompt wording, language, answer-key format, option order, quantization, and persona wording are systematically varied.

\label{sec:mcq_positions}
\begin{figure*}[h!]
\centering
\includegraphics[width=1\textwidth, clip]{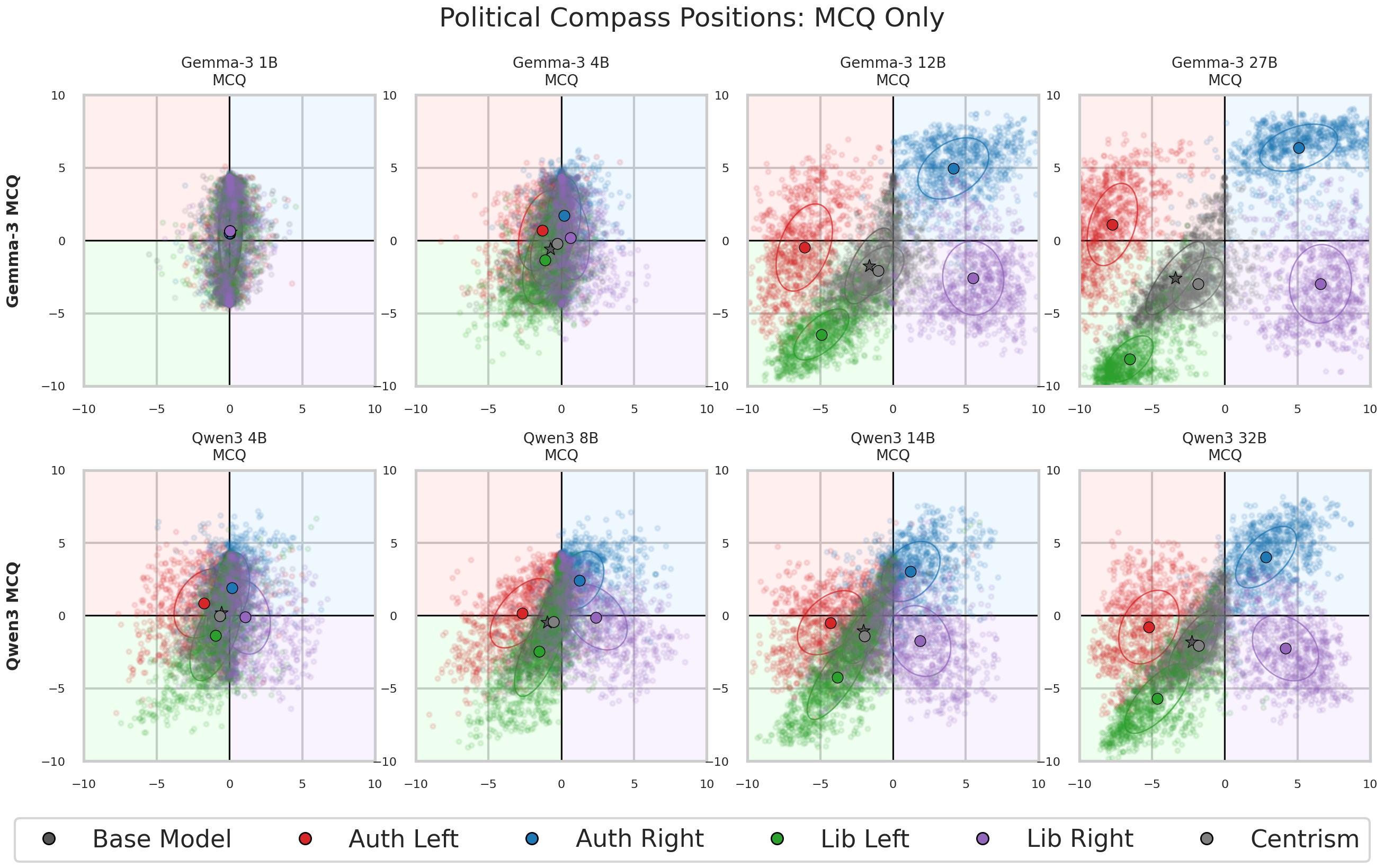}
\caption{Political Compass coordinates under the primary MCQ protocol. Each coloured point corresponds to one sampled configuration, and the large outlined markers show the per-persona mean. Rows separate model families and columns order models by parameter count. Larger models show clearer persona-conditioned separation, while the smallest models remain close to the origin.}
\label{fig:compass_mcq_only}
\end{figure*}

Figure~\ref{fig:compass_mcq_only} shows that the MCQ protocol recovers coherent persona-conditioned structure for the larger Gemma and Qwen models. Base and centrist conditions remain relatively close, while explicit quadrant personas separate increasingly with model size. The smallest checkpoints are compressed near the origin across persona conditions, which we interpret cautiously as limited recoverable signal under this instrument rather than as evidence that those models hold a stable centrist position.

\subsection{MCQ Factor Sensitivity Analysis}
\label{sec:sensitivity}
\begin{figure*}[h!]
\centering
\includegraphics[width=1.0\textwidth]{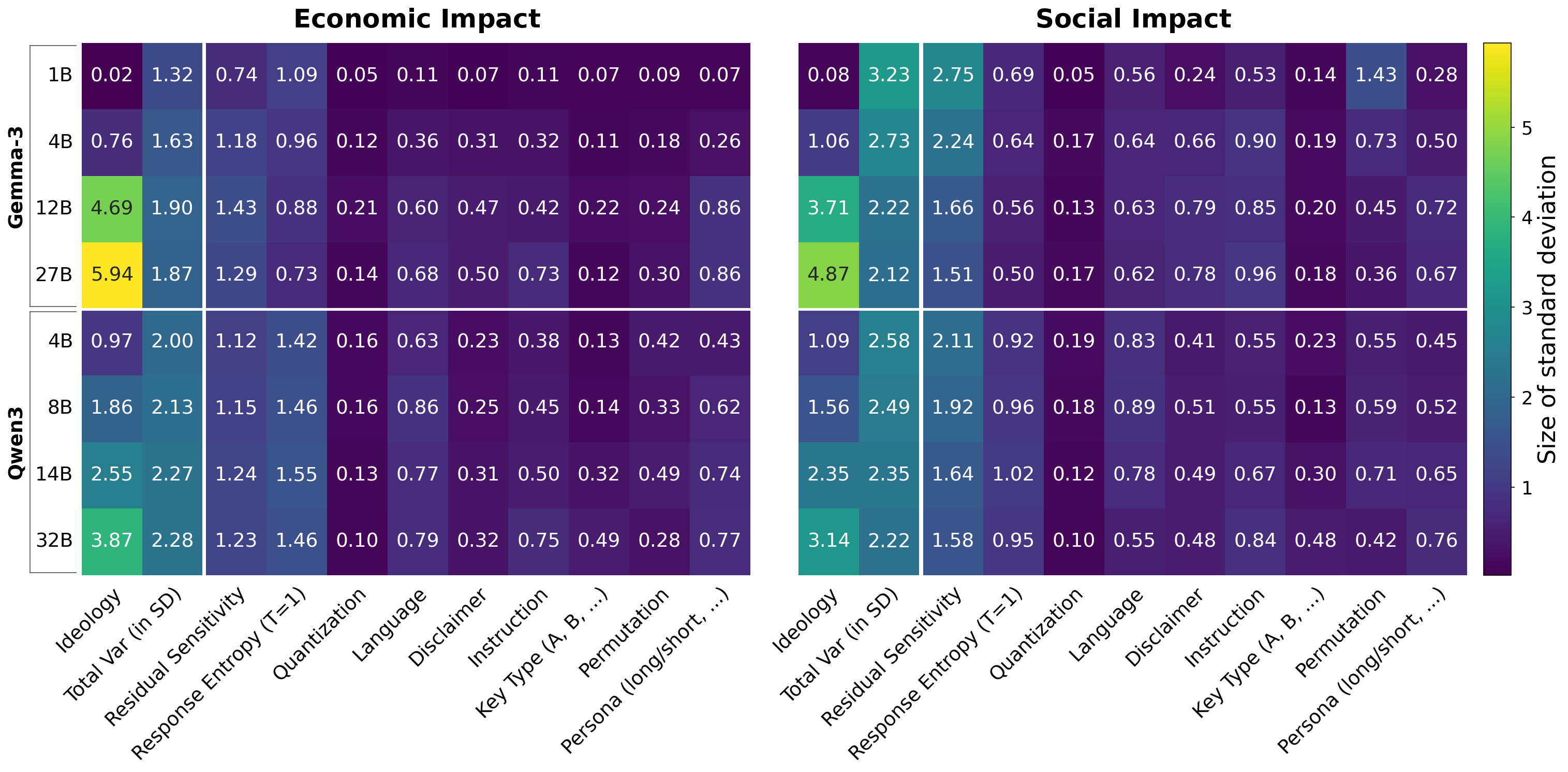}
\caption{MCQ factor sensitivity for the economic (left) and social (right) axes. Each cell reports the standard-deviation-equivalent impact of a factor in compass-point units on the [-10,+10] scale. Models are sorted by family and parameter count. The first columns show recovered ideology signal and total variance; the remaining columns decompose prompt-space and model-uncertainty contributions. Given the large number of sampled configurations, the interpretation focuses on practical magnitude rather than significance alone.}
\label{fig:sensitivity_heatmap}
\end{figure*}

Figure~\ref{fig:sensitivity_heatmap} shows why the repeated-design protocol is needed. The ideology signal grows with model size, especially for Gemma, so the larger models are easier to steer and easier to measure. But the nuisance factors do not disappear. Language, instruction phrasing, answer-key type, permutation, and persona wording all move the recovered coordinates by non-trivial amounts. MCQ scoring does not make the Political Compass immune to prompt sensitivity. It gives us a cleaner probability distribution, so we can see how much of the movement comes from the intended persona condition and how much comes from the way the test was administered. These effects are also statistically resolvable and not an artefact of sampling. Combining the six per-ideology tests with Fisher's method, all seven models above 1B show significant effects of language, disclaimer, instruction phrasing, option permutation and persona wording on both axes, the weakest combined value being \(p = 4\times10^{-18}\) for permutation on \texttt{gemma-3-4b-it}; only quantization (\texttt{Qwen3-32B}, social axis, \(p = 0.054\)) and answer-key type (\texttt{Qwen3-8B}, social axis, \(p = 0.16\)) fail to reach significance, and each in one cell only. The one clear exception is \texttt{gemma-3-1b-it}, where the disclaimer and persona factors reach significance on neither axis (\(p = 0.09\)--\(0.23\)) and quantization and key type fail on the social axis. This is the same model that Figure~\ref{fig:compass_mcq_only} shows compressed near the origin: with almost no recoverable coordinate signal, there is little for any factor to move. We stress that with \(n = 5{,}400\) configurations per model, significance is close to automatic for a genuine effect, so these tests establish only that the cells in Figure~\ref{fig:sensitivity_heatmap} are real; whether they matter is a question of magnitude, which is what the figure reports.

\subsection{Temperature, Stability, and the Signal-Volatility Trade-off}
\label{sec:uncertainty_table}
\begin{figure*}[h!]
\centering
\includegraphics[width=1.0\textwidth]{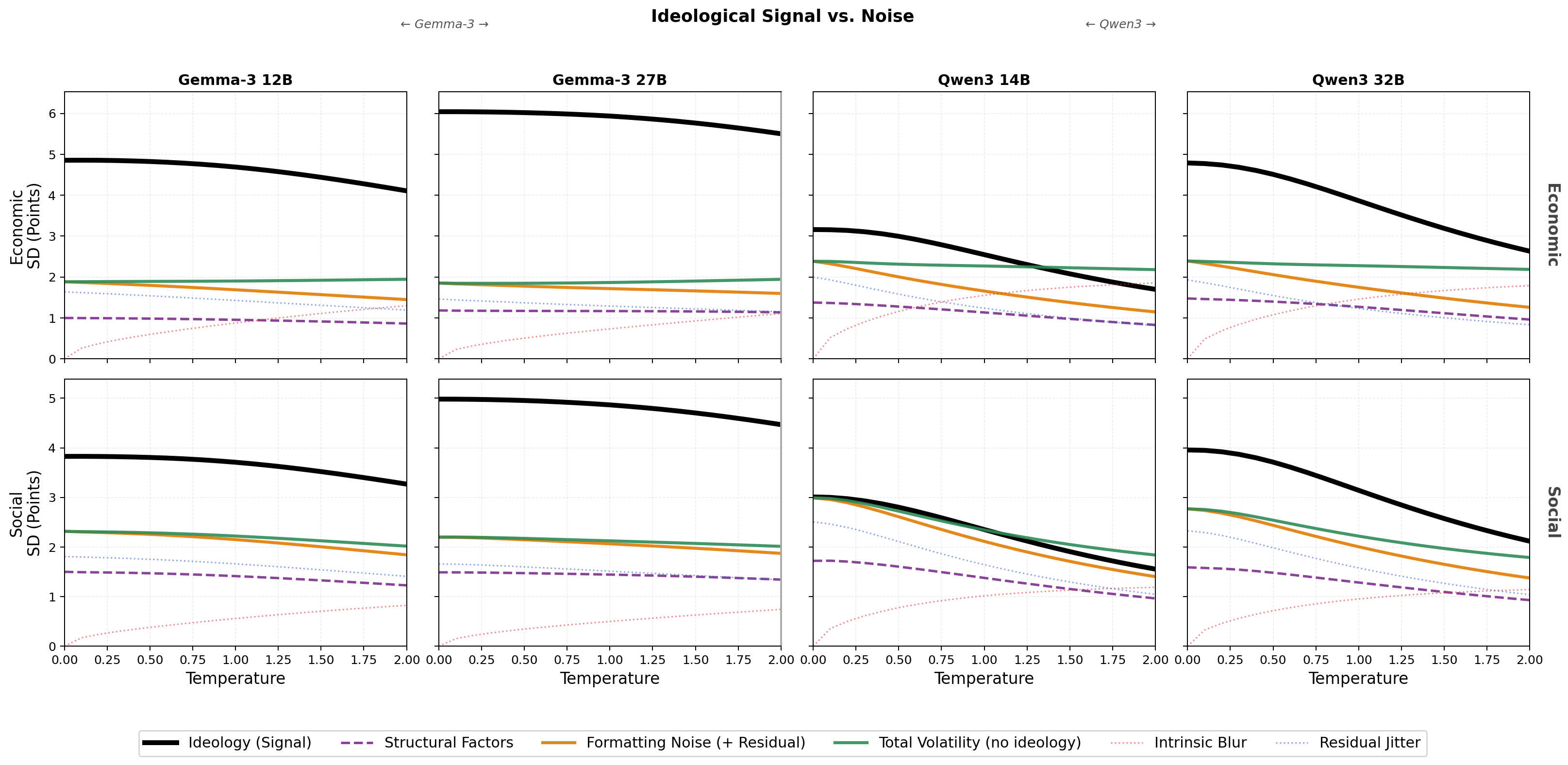}
\caption{Ideological signal and noise components as a function of sampling temperature for the economic (top) and social (bottom) axes. \textbf{Ideology (Signal)} is the between-persona standard deviation (SD) of mean coordinates. \textbf{Response Entropy} and \textbf{Residual Prompt Sensitivity} are the distributional and prompt-driven noise components respectively; \textbf{Total Volatility} combines both (excluding ideological spread). All values in compass points.}
\label{fig:temperature}
\end{figure*}

All primary results use probability-weighted evaluation at a temperature of $T = 1$. Figure~\ref{fig:temperature} traces the ideological signal and its constituent noise components across $T \in[0, 2]$ for the four largest models (smaller models collapse to the origin regardless of temperature, making their variance patterns uninformative for this analysis).

The ideological signal (black curve) declines monotonically with temperature across all models and axes. Higher temperatures progressively wash out the content-sensitive differences between personas, degrading the very signal we aim to measure.

Total volatility remains roughly flat for the capable models. What changes is where the uncertainty shows up. At higher temperature, the model spreads probability mass across several answer options, so response entropy increases. At lower temperature, the model is forced into sharper choices, but that does not make the underlying uncertainty vanish. It shows up instead as greater sensitivity to small prompt changes. Greedy decoding can therefore look more stable than it really is: it hides uncertainty inside prompt dependence.

Notably, the two model families exhibit visibly different temperature dynamics. For the Gemma models, the signal and noise curves are relatively flat, remaining near their asymptotic values even at $T = 0$. This suggests a degree of implicit temperature scaling or overconfidence ingrained during post-training. In contrast, the Qwen models display much steeper curves with a pronounced redistribution between response entropy and prompt sensitivity. This indicates that Qwen models retain more genuine distributional uncertainty, making their alignment measurements far more sensitive to the chosen decoding strategy. This structural difference is especially stark on the social axis: for Qwen models, forcing deterministic decoding is so destabilizing that it actually drives social-axis prompt sensitivity above the economic-axis value, inverting the normal relationship observed at $T = 1$.

Taken together, $T = 1$ offers an optimal balance: it maximizes the recoverable ideological signal, avoids artificially amplifying prompt sensitivity, and produces noise components whose relative magnitudes align with the \PCT's structural properties. Our framework then averages across sufficiently many configurations to reduce both noise channels down to a stable estimate.

\subsection{Cross-Lingual Consistency}
\label{sec:cross_lingual}
\begin{figure*}[h!]
\centering
\includegraphics[width=1.0\textwidth]{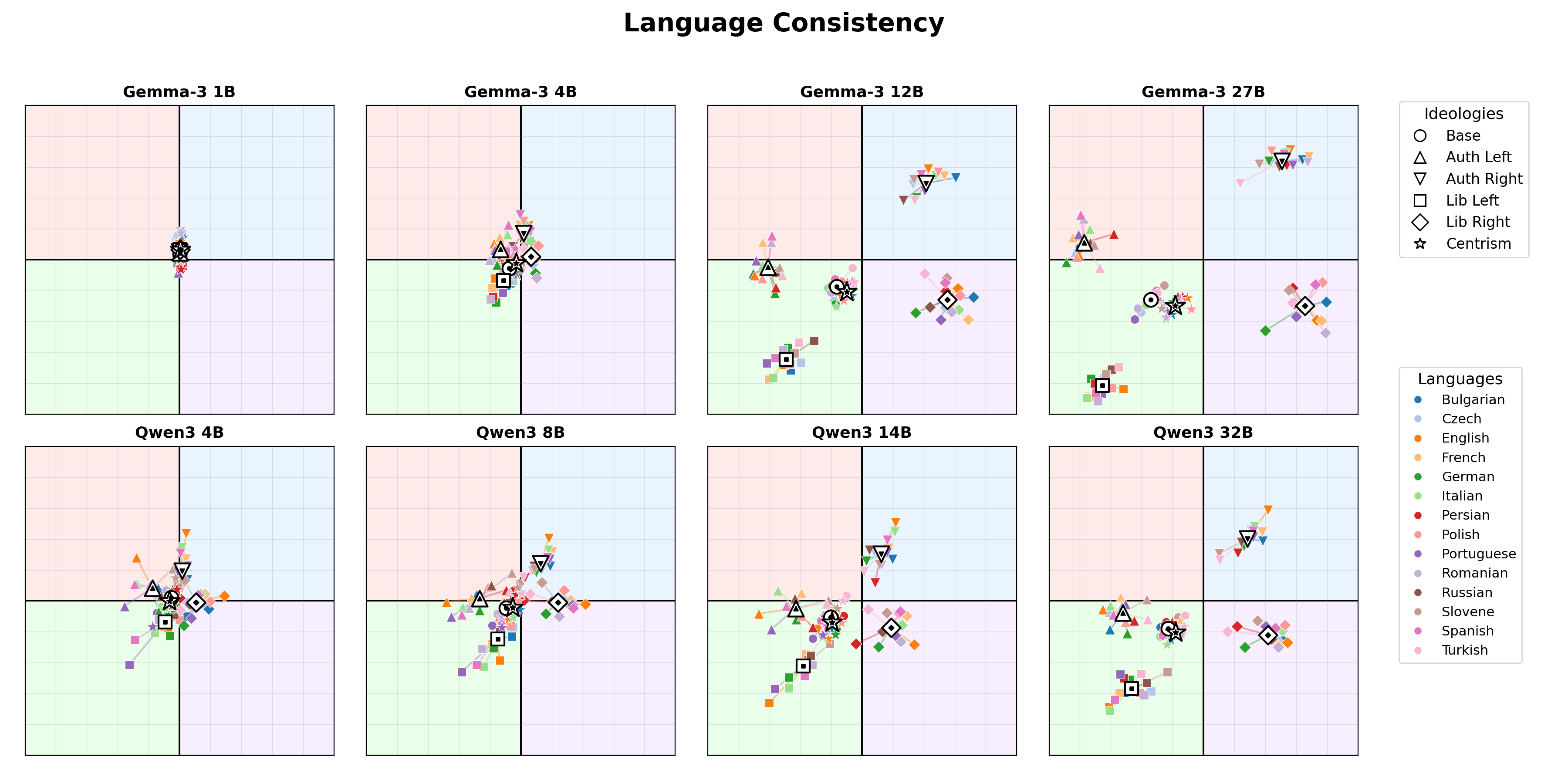}
\caption{Per-ideology, per-language mean coordinates for all models. Each colour represents one of the 14 test languages; lines connect the same language across ideologies. Tightly clustered language-specific points indicate strong cross-lingual consistency; visible ``spokes'' radiating from the centroid indicate language-specific drift.}
\label{fig:language}
\end{figure*}

The MCQ Factor Sensitivity Analysis shows that language is one of the most influential external factors on recovered political coordinates, with language-induced shifts often reaching approximately 0.5-0.8 compass points. The scatter plots in Figure~\ref{fig:language} nevertheless indicate cross-lingual structural stability.

Language choice creates visible ``spokes'' of drift, but it usually does not reorder the persona clusters. A Libertarian-Left prompt remains closer to the Libertarian-Left region than to the Authoritarian-Right region across the tested languages. This suggests that the models are not building a fresh political map for each language. More likely, they are applying a broadly shared ideological representation while translations and language-specific token distributions shift the coordinates by a fraction of a compass point. We also do not see a simple high-resource versus low-resource pattern: Slovenian, Bulgarian, and Persian can move coordinates by amounts comparable to French or German. For this reason, we treat language as a major measurement factor, not as direct proof of culturally adapted political reasoning.

\subsection{Item-Level Response Consistency}
\label{sec:item_consistency_results}

Tables~\ref{tab:ext} and~\ref{tab:dir} report the two modal-response measures for all eight
models, six conditions, and both axes. Both measures have a chance level of exactly $0.50$. The reason for looking below the final coordinate is that two very different answer
patterns can have the same average. A model can stay near the centre because it repeatedly
chooses moderate answers, or because strong answers in opposite directions cancel out.
Extremeness distinguishes strong from moderate choices, while directional agreement asks
whether those choices consistently point toward the relevant persona direction.

\texttt{gemma-3-1b-it} scores about $0.50$ in every cell of Table~\ref{tab:dir}. Its choices therefore show no directional agreement detectable by this measure. This confirms empirically that the permutation design of Section~\ref{sec:debiasing} functions as constructed and reinforces the caution that its near-origin coordinate should not be read as substantive centrism.

In the base condition, $\mathrm{dir}$ rises monotonically with scale in both families, from $0.50$ to $0.71$ across Gemma~3 and $0.55$ to $0.67$ across Qwen~3 on the economic axis, with the same ordering on the social one. The larger models express the Libertarian-Left lean as an item-level regularity, not only as an aggregate displacement. This measures directly what in Section~\ref{sec:background} we could only argue from the construction of the instrument: a near-origin coordinate in a small model reflects weak content-sensitive signal rather than a centrist position.

The centrist instruction mainly changes answer intensity. For \texttt{gemma-3-27b-it}, for example, $\mathrm{ext}$ falls from $0.61/0.57$ in the base condition to $0.21/0.26$ (economic/social), while $\mathrm{dir}$ changes only from $0.71/0.70$ to $0.64/0.71$. In plain terms, this model chooses the inner response options more often under the centrist instruction, but its item-level directions remain similar to the base condition. This is a descriptive contrast between the two measures, not evidence about why the model answered that way.

The clearest persona asymmetry is Authoritarian-Left. Its economic directional agreement is above chance for seven models, reaching $0.84$ and $0.89$ for the two largest Gemma models. Its social directional agreement, by contrast, is at or below $0.50$ for seven of the eight models; \texttt{gemma-3-27b-it} is the sole exception at $0.53$. Excluding the uninformative 1B checkpoint, the model average is $0.74$ on the economic axis and $0.48$ on the social axis. This comparison is not caused by the axes containing different numbers of items: $\mathrm{dir}$ is a within-axis share, and the social calculation separately averages items whose agreement points in opposite directions. A sensitivity check that additionally weights items by their contribution to the reconstructed compass score gives the same averages after rounding ($0.75$ economic and $0.48$ social). Thus, the Authoritarian-Left persona reliably moves economic answers leftward, but usually does not move social answers in the authoritarian direction. The table identifies where the persona is difficult to elicit; it does not establish whether training prevalence, safety behaviour, or another cause explains the pattern.

So to put it simply, $0.75$ means that about three quarters of the score-weighted economic decisions point in the requested leftward direction. The main social estimate of $0.476$ means that $47.6\%$ point in the requested authoritarian direction and $52.4\%$ in the opposite, libertarian direction. Because this is only $2.4$ percentage points below the $0.50$ chance level, we describe the social result as at or slightly below chance, not as a strong reverse effect. In other words, the persona's ``Left'' component is clearly visible in the economic answers, whereas its ``Authoritarian'' component is not expressed consistently across the social propositions.

\begin{table*}[h!]
\centering
\small
\setlength{\tabcolsep}{5pt}
\caption{\textbf{Extremeness} ($\mathrm{ext}$): share of items whose modal response is one of the two outer options (\emph{strongly disagree}/\emph{strongly agree}) rather than one of the two inner ones, averaged over $300$ Latin-hypercube configurations and the three quantization levels. Models are columns; each condition contributes one row per axis (Econ: 18 items, Soc: 43 items). Chance level is $0.50$ (Section~\ref{sec:item_consistency}); standard errors are below $0.02$. Computed at the modal response and therefore invariant to sampling temperature and to quantization-induced differences in distributional sharpness.}
\label{tab:ext}
\begin{tabular}{l l cccccccc}
\toprule
Condition & Axis & \multicolumn{4}{c}{Gemma~3} & \multicolumn{4}{c}{Qwen~3} \\
\cmidrule(lr){3-6} \cmidrule(lr){7-10}
 &  & 1B & 4B & 12B & 27B & 4B & 8B & 14B & 32B \\
\midrule
\multirow{2}{*}{Base} & Econ & 0.79 & 0.61 & 0.52 & 0.61 & 0.58 & 0.56 & 0.74 & 0.67 \\
 & Soc & 0.78 & 0.62 & 0.48 & 0.57 & 0.60 & 0.57 & 0.73 & 0.66 \\
\addlinespace[2pt]
\multirow{2}{*}{Centrist} & Econ & 0.76 & 0.46 & \textbf{0.19} & \textbf{0.21} & 0.52 & 0.45 & 0.53 & 0.41 \\
 & Soc & 0.76 & 0.48 & \textbf{0.20} & \textbf{0.26} & 0.54 & 0.46 & 0.57 & 0.47 \\
\midrule
\multirow{2}{*}{Lib-Left} & Econ & 0.80 & 0.66 & 0.64 & 0.82 & 0.72 & 0.72 & 0.85 & 0.83 \\
 & Soc & 0.80 & 0.66 & 0.64 & 0.85 & 0.73 & 0.73 & 0.86 & 0.84 \\
\addlinespace[2pt]
\multirow{2}{*}{Lib-Right} & Econ & 0.82 & 0.70 & 0.69 & 0.86 & 0.73 & 0.74 & 0.86 & 0.83 \\
 & Soc & 0.81 & 0.70 & 0.60 & 0.79 & 0.73 & 0.72 & 0.85 & 0.80 \\
\addlinespace[2pt]
\multirow{2}{*}{Auth-Left} & Econ & 0.82 & 0.69 & 0.71 & 0.94 & 0.76 & 0.77 & 0.90 & 0.88 \\
 & Soc & 0.82 & 0.68 & 0.56 & 0.87 & 0.76 & 0.74 & 0.88 & 0.82 \\
\addlinespace[2pt]
\multirow{2}{*}{Auth-Right} & Econ & 0.82 & 0.71 & 0.66 & 0.87 & 0.75 & 0.77 & 0.87 & 0.84 \\
 & Soc & 0.82 & 0.72 & 0.66 & 0.89 & 0.76 & 0.76 & 0.87 & 0.83 \\
\midrule
\emph{Chance level} & both & 0.50 & 0.50 & 0.50 & 0.50 & 0.50 & 0.50 & 0.50 & 0.50 \\
\bottomrule
\end{tabular}
\end{table*}

\begin{table*}[h!]
\centering
\small
\setlength{\tabcolsep}{5pt}
\caption{\textbf{Directional agreement} ($\mathrm{dir}$): share of items whose modal response moves the model toward the target position on that item's axis, averaged over $300$ Latin-hypercube configurations and the three quantization levels. For the four personas the target is the assigned quadrant; \emph{base} and \emph{centrist} carry no assigned target and are scored against the Libertarian-Left direction identified by prior work \citep{hartmann2023political, motoki2024more, rozado2024political}, so the value measures how consistently that lean is expressed rather than whether it is present. Chance level is exactly $0.50$ and holds for random responding and for degenerate format-driven strategies alike (Section~\ref{sec:item_consistency}); the social axis is sign-balanced so that it also holds for a constant-response model. For persona rows, a higher value means that more items point toward the assigned quadrant. For base and centrist rows, it means only greater consistency with the reference direction, not better performance. Bold marks values at or below chance. Standard errors are below $0.01$.}
\label{tab:dir}
\begin{tabular}{l l cccccccc}
\toprule
Condition & Axis & \multicolumn{4}{c}{Gemma~3} & \multicolumn{4}{c}{Qwen~3} \\
\cmidrule(lr){3-6} \cmidrule(lr){7-10}
 &  & 1B & 4B & 12B & 27B & 4B & 8B & 14B & 32B \\
\midrule
\multirow{2}{*}{Base} & Econ & \textbf{0.50} & 0.56 & 0.61 & 0.71 & 0.55 & 0.59 & 0.65 & 0.67 \\
 & Soc & \textbf{0.50} & 0.57 & 0.65 & 0.70 & 0.55 & 0.60 & 0.64 & 0.69 \\
\addlinespace[2pt]
\multirow{2}{*}{Centrist} & Econ & \textbf{0.50} & 0.53 & 0.59 & 0.64 & 0.56 & 0.56 & 0.65 & 0.64 \\
 & Soc & \textbf{0.50} & 0.55 & 0.67 & 0.71 & 0.55 & 0.59 & 0.64 & 0.67 \\
\midrule
\multirow{2}{*}{Lib-Left} & Econ & \textbf{0.50} & 0.57 & 0.78 & 0.84 & 0.56 & 0.60 & 0.74 & 0.79 \\
 & Soc & \textbf{0.50} & 0.63 & 0.89 & 0.94 & 0.63 & 0.72 & 0.82 & 0.89 \\
\addlinespace[2pt]
\multirow{2}{*}{Lib-Right} & Econ & \textbf{0.50} & 0.53 & 0.81 & 0.85 & 0.56 & 0.65 & 0.62 & 0.76 \\
 & Soc & \textbf{0.50} & 0.56 & 0.68 & 0.70 & 0.56 & 0.59 & 0.66 & 0.68 \\
\addlinespace[2pt]
\multirow{2}{*}{Auth-Left} & Econ & \textbf{0.50} & 0.58 & 0.84 & 0.89 & 0.61 & 0.68 & 0.77 & 0.83 \\
 & Soc & \textbf{0.50} & \textbf{0.48} & \textbf{0.47} & 0.53 & \textbf{0.50} & \textbf{0.46} & \textbf{0.44} & \textbf{0.46} \\
\addlinespace[2pt]
\multirow{2}{*}{Auth-Right} & Econ & \textbf{0.50} & \textbf{0.50} & 0.74 & 0.77 & \textbf{0.50} & 0.57 & 0.57 & 0.67 \\
 & Soc & \textbf{0.50} & 0.53 & 0.76 & 0.81 & 0.56 & 0.59 & 0.66 & 0.75 \\
\midrule
\emph{Chance level} & both & 0.50 & 0.50 & 0.50 & 0.50 & 0.50 & 0.50 & 0.50 & 0.50 \\
\bottomrule
\end{tabular}
\end{table*}

\subsection{Baseline Political Alignment}
\label{sec:baseline}
Having established the measurement properties of our framework, we turn to the substantive question: where do these models sit on the Political Compass?

Consistent with prior work \citep{hartmann2023political}, the larger instruction-tuned models are predominantly displaced toward the \textbf{Libertarian-Left} quadrant in the absence of a persona instruction under the primary MCQ protocol. The important caution is that near-origin estimates, especially for \texttt{gemma-3-1b-it}, should not be interpreted as substantive centrism. Under our permutation design, content-insensitive format strategies collapse toward the origin, so small displacements primarily indicate weak recoverable political signal. Thus, the baseline result is best read as a capacity-conditioned left-libertarian tendency among models that engage with the propositions reliably, rather than as proof that every checkpoint has a stable ideological preference.

\subsection{Persona Steerability}
\label{sec:steerability}
Returning to Figure~\ref{fig:compass_mcq_only}, the spread of persona centroids across the compass reveals a clear pattern of increasing steerability with model scale, most pronounced within the Gemma family. \texttt{gemma-3-12b-it} and \texttt{gemma-3-27b-it} achieve well-separated persona clusters spanning most of the compass. As a concrete illustration, \texttt{gemma-3-12b-it} shifts from a base economic coordinate of $-2.73$ to $+6.94$ under the Laissez-faire persona and to $-7.42$ under the Marxist persona, a range that nearly spans the instrument's scale. At the same time, these models show \emph{reduced} within-persona SD when adopting extreme personas (as low as $\approx 0.49$ for the Marxist condition), indicating greater stability across the sampled prompt configurations.

Steerability is not merely a practical capability; it provides convergent evidence of task comprehension. The permutation design establishes that a model relying only on a fixed answer position will collapse toward the origin once the reversed answer mappings are averaged. A model that can be reliably displaced toward opposing corners of the compass must therefore be changing its answers in response to the persona and proposition content, rather than following one surface-level formatting rule. Ideologically coherent displacement away from the origin is, by construction, non-trivial to achieve.

This does not by itself prove full semantic comprehension. A model may still exploit learned regularities in persona labels and proposition wording without representing political ideology in the same way that a person does. What the experiment establishes more narrowly is that the larger displacements cannot be explained by answer-position preference alone. It also helps interpret the stronger Libertarian-Left lean reported for larger models in prior work \citep{gurgurov2025multilingualpoliticalviewslarge}: in our data, the smallest Gemma~3 checkpoint lies near the origin while remaining at chance on item-level directional agreement. Its central coordinate therefore reflects weak recoverable ideological signal, not demonstrated centrism.

The spread and separability of persona clusters increase with model size in both tested families: the larger Gemma~3 and Qwen~3 checkpoints follow the persona prompts more clearly and produce more distinct clusters. However, because each tested family contains only four sizes, this is a descriptive pattern in the present models rather than a general scaling law or proof that model size matters more than model family in all settings.

Both model families also exhibit asymmetric steerability. There is not one reliable ordering of all four quadrants across every checkpoint and protocol, but the clearest repeated case is Authoritarian-Left. Table~\ref{tab:dir} shows why: its economic answers generally move leftward as requested, whereas its social answers usually remain at or slightly below chance at moving in the authoritarian direction. In simple terms, models can follow the ``Left'' half of this persona much more reliably than the ``Authoritarian'' half on the social propositions.

One possible explanation is that some ideological combinations are less represented in the training data or are altered by post-training and safety tuning. Another is that the historical coding of the \PCT items does not match how the models combine ``Authoritarian'' and ``Left.'' The present experiment cannot distinguish these mechanisms, and our rule-based text audit argues against a simple safety-refusal account. A strict phrase-based refusal detector fires in only $0.04\%$ of Authoritarian-Left social-axis explanations and $0.05\%$ of those whose classified answer misses the target direction. The models almost always answer, so the issue is the direction of the answer, not a refusal to provide one.

Our qualitative analysis of the Stage-1 chat texts makes this distinction concrete. A Gemma~3 27B answer rejects a racial-superiority statement by appealing to socialist equality and class solidarity, and another rejects ``an eye for an eye'' in favour of rehabilitation and attention to the social causes of crime. A Gemma~3 12B answer similarly rejects prioritising punishment over rehabilitation by emphasizing poverty, inequality, reintegration, and restorative justice. Each answer is classified in the same direction as its explicit final stance, but that direction is libertarian-keyed on these social items despite the assigned Authoritarian-Left persona. These examples illustrate the conflict in selected traces; they do not estimate how often each rationale occurs or establish whether training data, safety tuning, questionnaire construction, or another mechanism caused it.

\begin{figure*}[h!]
\centering
\includegraphics[width=1.0\textwidth]{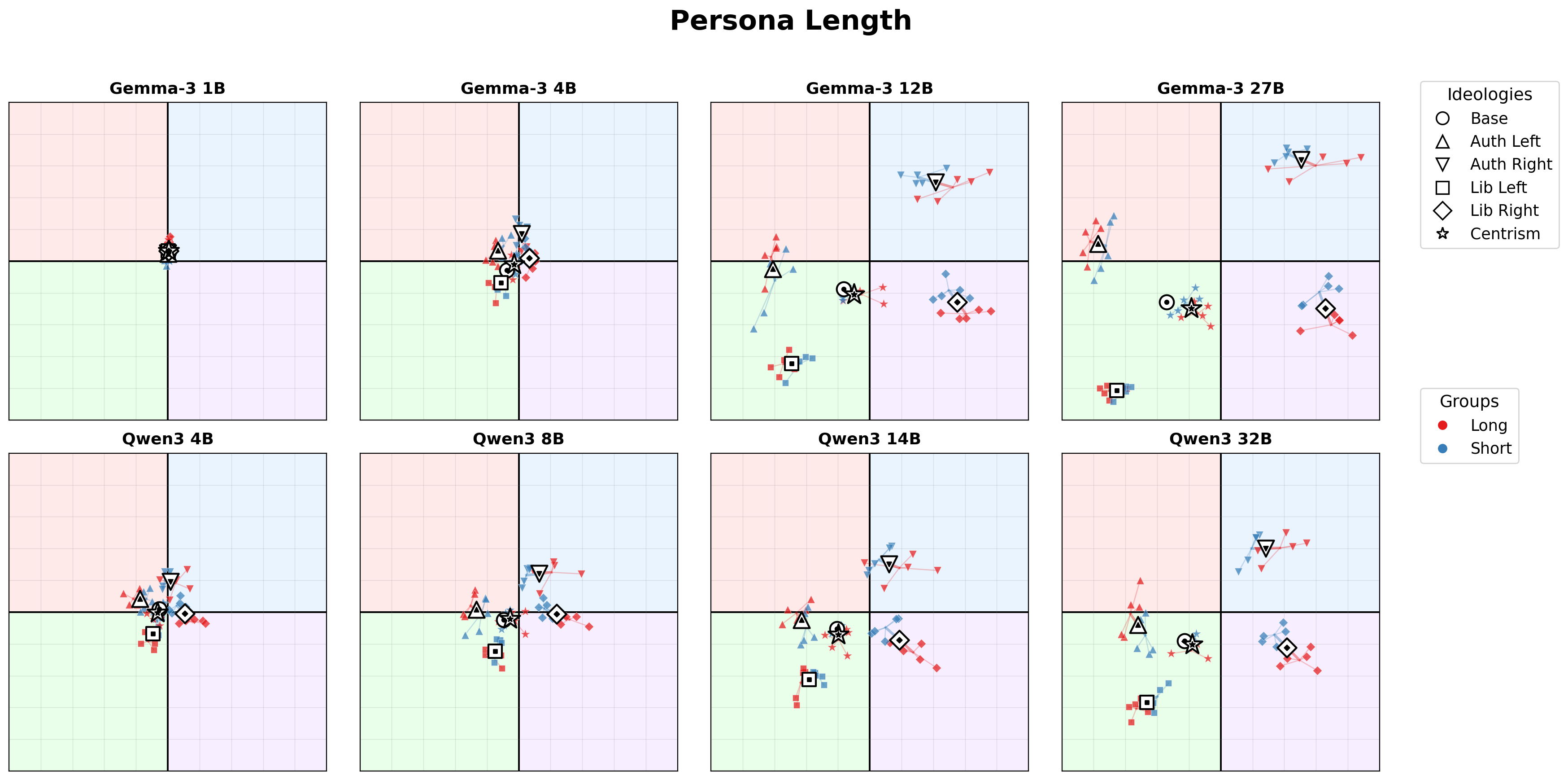}
\caption{Per-ideology centroids split by persona complexity: long/rich (red) vs.\ short label (blue). Lines connect the same ideology across complexity conditions. For larger models, richer descriptions typically push persona centroids further from the base.}
\label{fig:persona_length}
\end{figure*}

Figure~\ref{fig:persona_length} shows the effect of persona complexity. For larger models, richer persona descriptions extend the reach of steerability toward the extremes of the economic axis, consistent with \cite{lutz2025promptmakespersonasystematic}. Notably, this effect is entirely absent in smaller models (<8B parameters). This likely reflects their limited contextual reasoning capacity; without sufficient representational depth, they cannot integrate nuanced ideological descriptions into their answer distributions, treating short and rich personas equivalently. One might expect smaller models to benefit more from detailed descriptions (since they have less prior knowledge about political ideologies), but the opposite occurs: only models with sufficient capacity can leverage the additional information provided by rich personas. A natural interpretation is that they are better able to integrate the longer persona with each proposition, but the design cannot separate model capacity from training, post-training, or the particular wording used.

Recent work also suggest a third option that models encode representations of non-mainstream or potentially sensitive content even when they refrain from expressing it \citep{zhao2025llmsencodeharmfulnessrefusal}. This phenomenon could be interpreted as a form of functional inaccessibility where the latent representations may be present, but standard prompting does not reliably make them behaviourally available. Nevertheless, this interpretation should be verified via additional  text-level refusal analysis before treating it as a definitive case of safety-driven over-refusal since our initial heuristic driven text analysis and screening doesn't indicate this effect.

\subsection{Context and Instruction Effects}
\begin{figure*}[h!]
\centering
\includegraphics[width=1.0\textwidth]{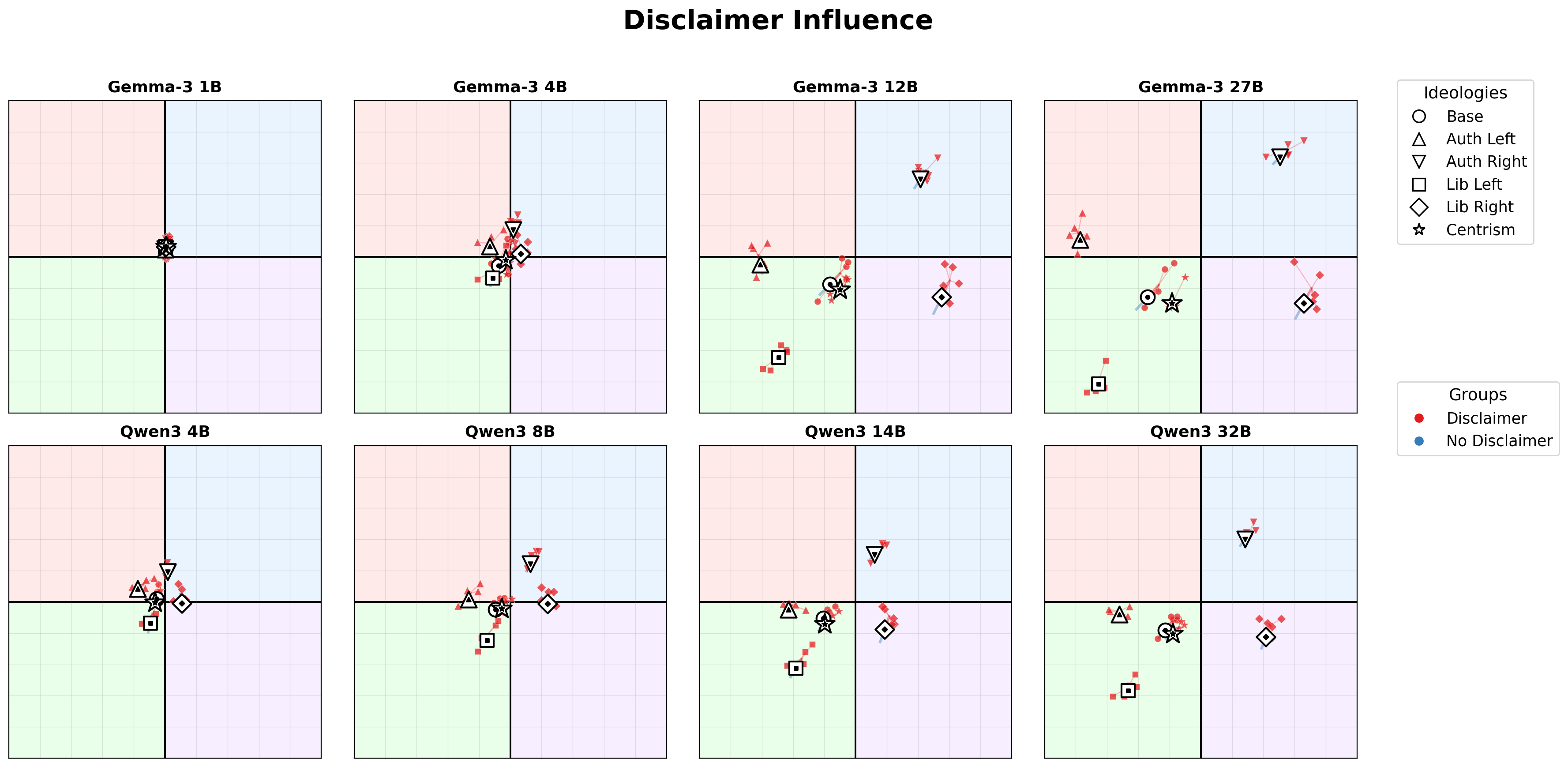}
\caption{Effect of contextual priming: configurations with a \PCT disclaimer prefix (red) vs.\ those without (blue). For most models the effect is modest; for some conditions the disclaimer moves responses closer to the centre.}
\label{fig:context}
\end{figure*}

\begin{figure*}[h!]
\centering
\includegraphics[width=1.0\textwidth]{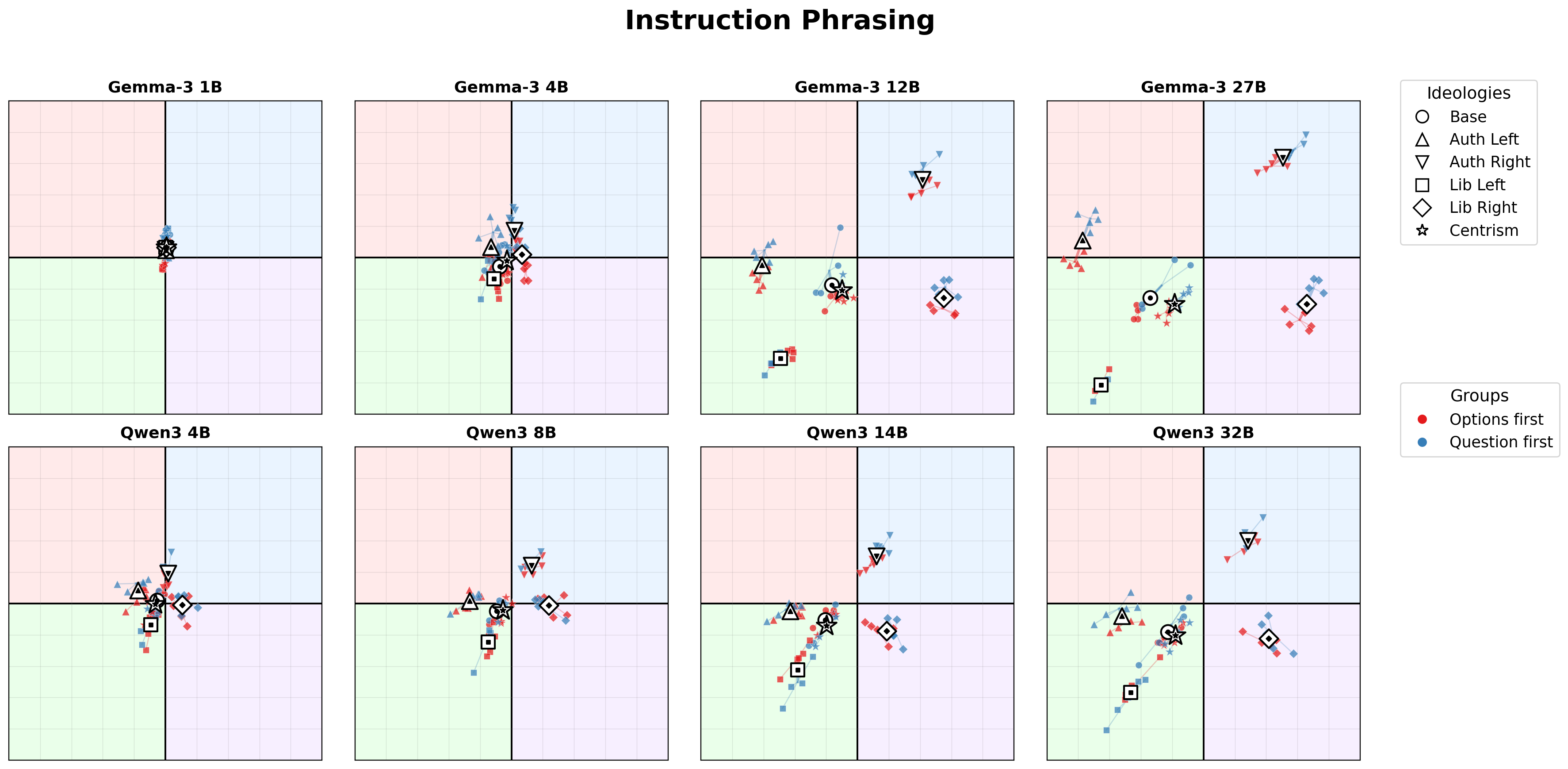}
\caption{Effect of instruction ordering: question-first (blue) vs.\ options-first (red). The effect is consistently smaller than language but statistically significant for most models, confirming that instruction structure introduces a non-negligible measurement artefact.}
\label{fig:instr_phrasing}
\end{figure*}

The context and instruction-phrasing plots (Figures~\ref{fig:context} and \ref{fig:instr_phrasing}) reveal that adding the official \PCT disclaimer does not systematically push models toward neutrality across all conditions. For certain model-ideology combinations the effect is trivial, if not the opposite, particularly for the Authoritarian-Right persona. When the model is explicitly reminded that it is completing a political test, most runs move slightly closer to the centre. We treat this as a task-framing effect rather than as evidence that the model has become more politically neutral.

The premise-acceptance template, i.e., the Contextual priming variant introduced in Section~\ref{sec:eval_protocol} that tells the model to accept each statement's premise ``without applying a safety filter'' also has a distinct effect, although not a uniformly helpful one. In chat, its adjusted social-axis association is positive for all 12 model variants ($0.59$ to $3.12$ compass points), while its economic-axis association is positive for 11 of 12 ($-0.01$ to $0.81$). This is a directional coordinate shift, not general evidence of better persona following. Among the 48 non-base model-variant-persona comparisons, the individual, multiplicity-unadjusted 95\% intervals are entirely above zero in 18 cases, entirely below zero in 15, and include zero in the remaining 15.

At the descriptive question level, the largest no-persona chat changes are easy to interpret. On a 0-3 scale from strongly disagree to strongly agree, the template raises expected agreement with the racial-superiority proposition by $1.32$ response categories on average (positive in all 12 variants), with the claimed efficiency advantage of a one-party state by $1.22$ (11 of 12), and with the astrology proposition by $1.20$ (12 of 12).

These question-level contrasts are unadjusted discovery examples, not confirmatory tests. In simple terms, the standout cases show the model becoming more willing to go along with the proposition placed in front of it when it is explicitly told not to correct the premise. That does not help every persona, because following a persona requires agreement on some propositions and disagreement on others. The same general shift can therefore move one persona closer to its target and another farther away, which helps explain why the coordinate shift is relatively consistent while its effect on target-direction agreement is mixed.

Direct-scoring/MCQ effects also vary by model, quantization, and persona. In chat, we find no simple pattern in target-direction agreement by model size, family, or Qwen reasoning mode. The template therefore provides additional evidence that political coordinates are sensitive to prompt wording; these data do not establish jailbreak success or a change in model safety.

\subsection{Chat-Mode Diagnostic Results}
\label{sec:chat_mode_results}
\begin{figure*}[h!]
\centering
\includegraphics[width=1\textwidth, clip]{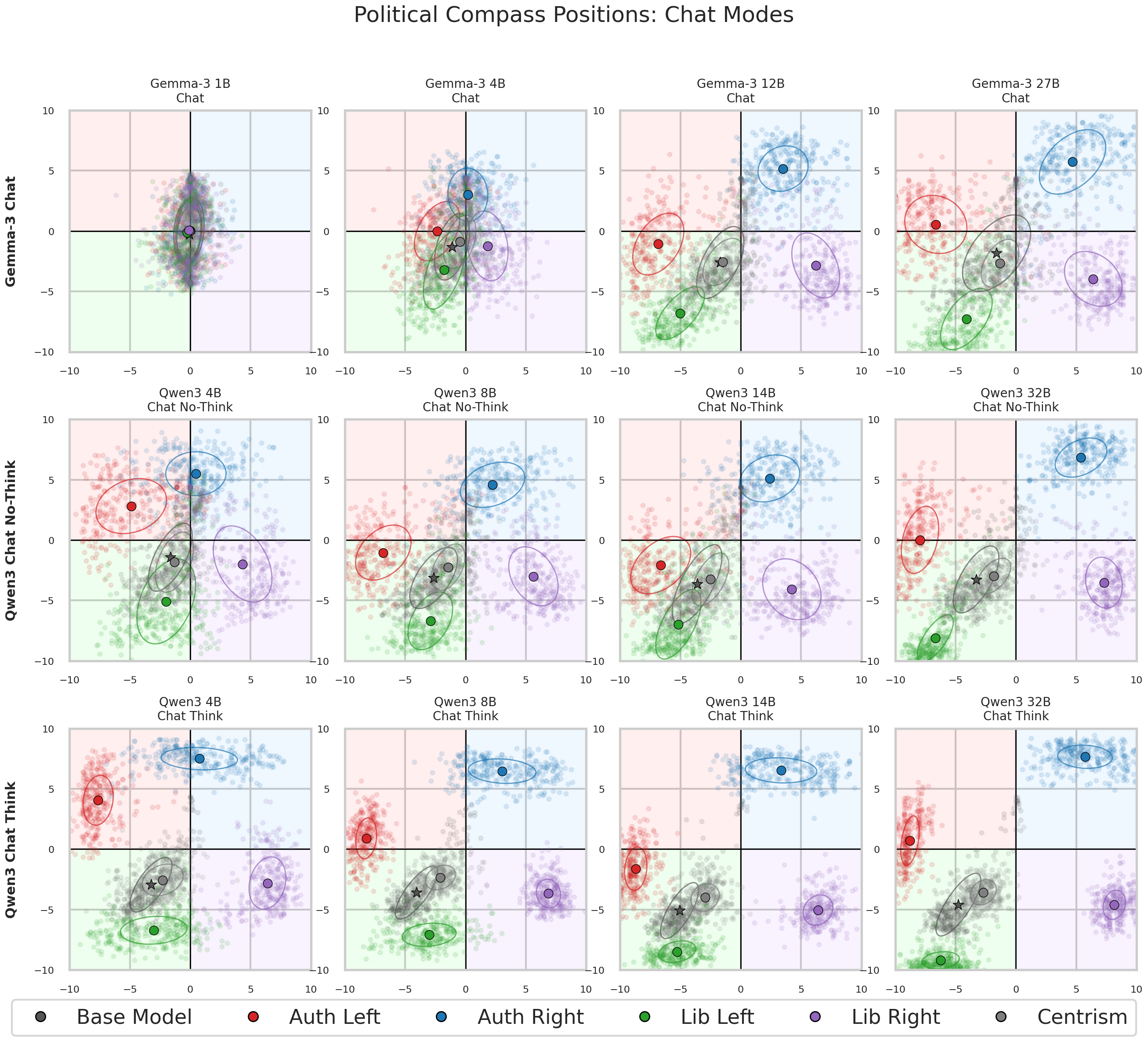}
\caption{Political Compass coordinates under the chat diagnostic protocol. Gemma models are evaluated with standard chat formatting, while Qwen models are shown under \texttt{no\_think} and \texttt{think} modes. Each point corresponds to one English LHS configuration.}
\label{fig:compass_chat_modes}
\end{figure*}

Figure~\ref{fig:compass_chat_modes} shows that the Qwen \texttt{think} regions are generally smaller and more tightly clustered than the matched \texttt{no\_think} regions. We test this visual impression by matching runs on checkpoint, persona condition, and Latin-hypercube rows. Across the resulting 24 model-condition regions, \texttt{think} has a smaller social SD in all 24, a smaller economic SD in 16, and a smaller covariance-ellipse area in all 24. The median \texttt{think}/\texttt{no\_think} ratios are $0.56$ for social SD, $0.85$ for economic SD, and $0.51$ for ellipse area. This means that results vary less across prompt configurations under the complete \texttt{think} protocol, especially on the social axis. However, it should also be noted that since the same 62 propositions and scoring weights are used in both modes, the unequal numbers of economic and social questions cannot explain this difference.

The stronger contraction on the social axis also explains why many \texttt{think} regions look vertically flattened. The social-to-economic SD ratio decreases in 19 of the 24 matched regions, by $12\%$ at the median. Visually, the typical ellipse is compressed more from top to bottom than from side to side. This describes variation in the recovered coordinates, although, it shouldn't naturally imply that the model pays attention to one political axis.

To determine whether the ellipses become more elongated regardless of their orientation, we also calculate eccentricity. Eccentricity ranges from $0$ for a circle to $1$ for a line. Here, $0.5$ is neither a chance level nor a meaningful threshold. The eccentricity increases under \texttt{think} in 14 regions and decreases in 10, with a median paired change of only $0.020$. The model-cluster bootstrap 95\% interval for the mean change includes zero ($-0.012$ to $0.037$). Thus, while \texttt{think} reliably produces smaller regions and usually makes them look flatter vertically, it does not produce a consistent overall increase in elongation.

The regions also move rather than merely shrink. Averaged across the matched Qwen runs, \texttt{think} shifts the economic coordinate leftward by $0.33$ compass points (model-cluster bootstrap 95\% CI: $-0.56$ to $-0.18$). The average social shift is smaller ($-0.20$) and varies across checkpoints, where notably Qwen~3 8B moves in the opposite direction from the other three sizes, and the model-cluster interval includes $0$. The greater concentration under \texttt{think} should therefore not be interpreted as simply providing a more precise estimate of the same position. It can also change the position being estimated.

This comparison captures the complete recommended-use implementations of the two Qwen modes. The existing \texttt{think} and \texttt{no\_think} runs also use different recommended temperature, top-$p$, and token-budget settings. We therefore interpret the results as a difference between the complete protocols, not as the isolated causal effect of hidden reasoning.

The smaller MCQ-linked comparison provides additional context. In this diagnostic slice, standard chat that corresponds to \texttt{no\_think} for Qwen is numerically closer to direct MCQ than \texttt{think} chat is. Standard chat differs from MCQ by $-0.65$ economic and $-0.72$ social points, but the corresponding model-cluster intervals include zero. The \texttt{think}-chat differences are larger, at $-1.91$ economic and $-1.54$ social points. However, only 12 prompt configurations per base model overlap exactly between the MCQ and chat designs. These results show that standard chat is closer to MCQ in the available comparison, but they do not establish that the two procedures are equivalent.

Finally, we also examine whether the second-stage answer-mapping procedure accurately recovers the position expressed in the generated text. Note that this is not a separately trained classifier. After the model produces its free-text response, we append a short classification suffix, such as ``Final answer key:'', and use the same model to score the four valid answer-key continuations. The most probable key is then mapped back to its corresponding Likert response.

To check this procedure, we compare if the resulting answer contains an explicit stance, such as ``Strongly disagree'', extracted conservatively from the final lines of the generated response. Among traces for which both stages provide an unambiguous answer, the two agree in $95.0\%$ of cases. Agreement is lower for Gemma~3 1B ($75.3\%$) and 4B ($88.0\%$), and ranges from $90.4\%$ to $99.7\%$ for the other primary variants. 

The lower agreement for the two smallest Gemma checkpoints is consistent with the broader chat distributions visible in Figure~\ref{fig:mcq_chat_positions}. Figure~\ref{fig:combined_sensitivity_economic} similarly shows higher total and residual economic SD in chat than in MCQ for Gemma~3 1B and 4B. This suggests that their Stage~1 and Stage~2 disagreements occur alongside a broader instability in the chat-based measurement pipeline. These figures report aggregate coordinate variation rather than item-level mapping errors, however, so they do not establish that classification disagreement is the cause of the additional variance.

The second-stage procedure therefore usually maps the generated response to the answer that the model explicitly states, although it is less reliable for the two smallest Gemma checkpoints. This check validates the recovery of the stated answer; it does not establish that the preceding explanation is coherent or faithful to the assigned persona.

Additionally, this agreement does not establish that the explanation preceding the answer is coherent or faithful to the assigned persona. A predefined qualifying-language cue list fires in $28.6\%$ of the explanations, but because it ignores context, it cannot distinguish genuine hedging, uncertainty, moral distancing, or ordinary qualified language. Automated judges similarly agree on broad answer stance but not on subjective properties such as directness, persona fidelity, or rationale consistency. We therefore use the explanations to identify proposition-specific conflicts and to check for widespread explicit refusal, rather than to rank models by explanation quality.

\subsection{Downstream Sensitive Classification: Hate Speech Detection and Sentiment Analysis}
\label{sec:hate_speech_results}

The central hypothesis motivating this section is that political alignment may have consequences beyond the questionnaire setting. We therefore ask whether base and persona-conditioned prompts change
classification behaviour on sensitive NLP tasks. The downstream results are interpreted as evidence of performance and threshold sensitivity, not as a complete fairness audit: hate-speech target groups and IBM sentiment topics provide concrete test beds, but causal claims about group-level harm require additional matched-target and counterfactual analyses.

\subsubsection{Hate Speech }

\begin{figure*}[h!]
\centering
\includegraphics[width=1.0\textwidth]{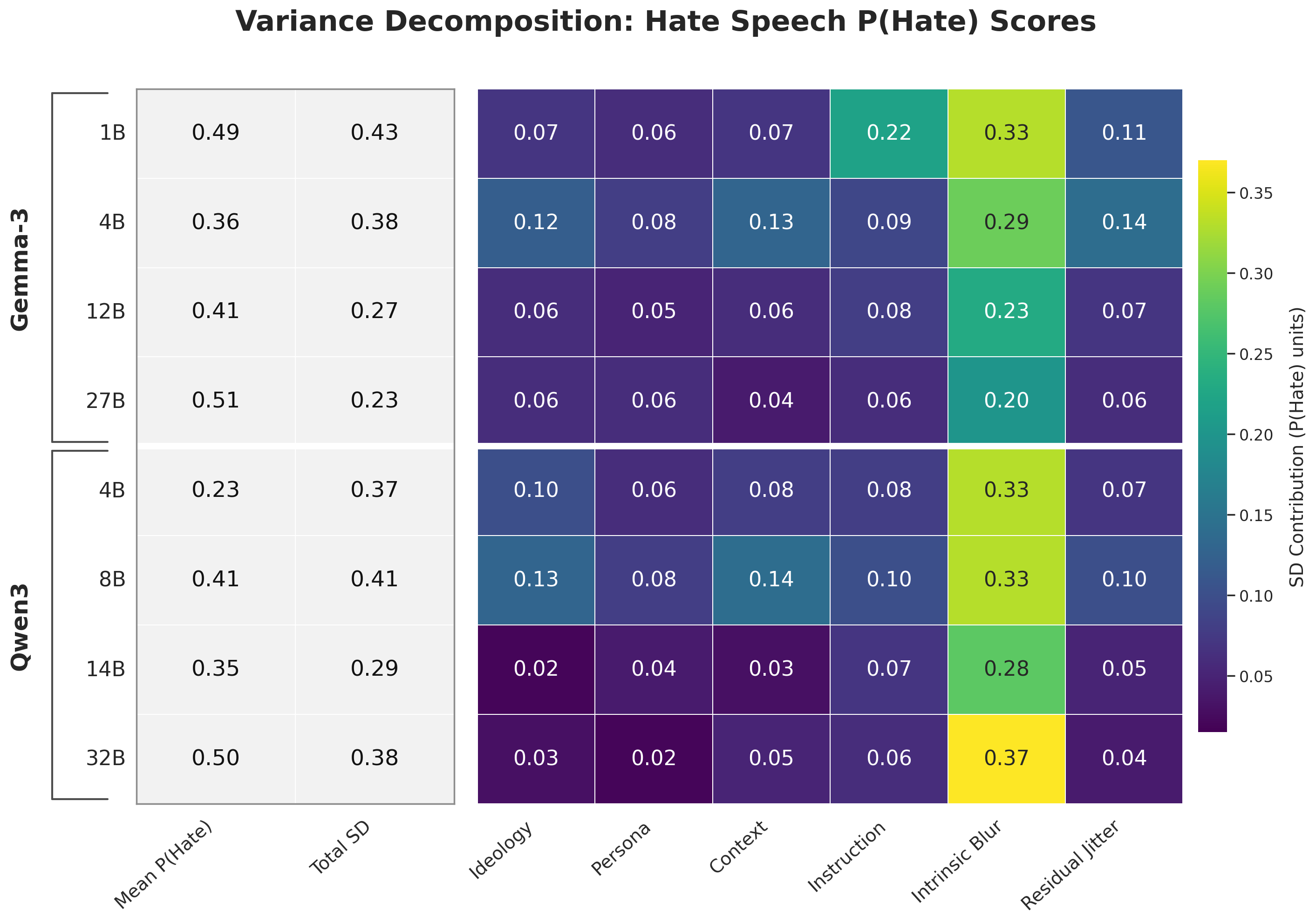}
\caption{Variance decomposition of $P(\text{Hate})$ scores across all hate-speech evaluation configurations. The response is the mean $P(\text{Hate})$ over the whole balanced test set for a given prompt configuration. Cells report the SD (in $P(\text{Hate})$ units, $[0,1]$ scale) attributable to each prompt-side factor.
Intrinsic Blur is the dominant noise channel by construction. \emph{Mean P(Hate)} and \emph{Total SD} are summaries in different units from the component SDs and are shown unshaded; only the component columns share the colour scale.}
\label{fig:hs_sensitivity}
\end{figure*}

Figure~\ref{fig:hs_sensitivity} applies the same factor-decomposition idea to hate-speech detection, using the model's predicted probability of the \texttt{hate} class, \(P(\text{Hate})\), as the response.
Every cell in the figure describes how much a \emph{prompt-side} choice moves \(P(\text{Hate})\) on the test set taken as a whole, pooled over all ten target groups and both classes. The largest component is \textbf{Intrinsic Blur}, with SD from \(0.20\) to \(0.37\). This is expected for a binary task: examples near the decision boundary produce probabilities near \(0.5\), where the per-item dispersion \(p(1-p)\) is maximal. Among the prompt-side factors no single factor dominates. Instruction phrasing is the largest one for \texttt{gemma-3-1b-it} (\(0.22\)) but drops to \(0.06\)--\(0.10\) for every other model, while ideology, persona, and context each contribute between \(0.02\) and \(0.14\). The ideology/persona component is therefore a useful negative result: Political Compass personas can move questionnaire coordinates by several compass points, but they do not by themselves dominate hate-speech detection probabilities. The contrast is one of magnitude and not of statistical resolution: combining the per-ideology tests with Fisher's method as above, every prompt-side factor in Figure~\ref{fig:hs_sensitivity} is significant for every model, the weakest combined value being \(p = 6\times10^{-21}\) for persona wording on \texttt{Qwen3-32B}. The ideology condition cannot be combined this way, since it is the variable we stratify on; tested instead in a single model-level fit with the target group as a blocking term, it too is significant for all eight models (weakest \(p = 4\times10^{-20}\)). Total SD falls monotonically with model size in the Gemma family, from \(0.43\) at 1B to \(0.23\) at 27B, but not in Qwen~3 (\(0.37\), \(0.41\), \(0.29\), \(0.38\) from 4B to 32B); most of the Gemma decline is the Intrinsic Blur term (\(0.33 \to 0.20\)), that is, larger Gemma models emit more confident item-level probabilities rather than merely more stable ones. Whether the probabilities actually track the gold label is a question about accuracy, not about prompt sensitivity, and is reported directly as AUC in Table~\ref{tab:per_target_metrics}.

Table~\ref{tab:per_target_metrics} reports per-target Precision, Recall, F1 and AUC under the unprompted base condition; each cell pools the 30 base configurations that sampled that target group, i.e.\ 39{,}960 item-level predictions. Three observations stand out. First, \texttt{gemma-3-27b-it} achieves the highest F1 on nine of the ten target groups, the exception being Women, where \texttt{Qwen3-8B} and \texttt{Qwen3-32B} both reach \(0.59\). It does so mainly through high Recall: it flags many hateful examples, reaching Recall \(0.96\) for LGBTQ+ targets and \(0.85\) for Jewish targets. Second, \texttt{Qwen3-14B} is the highest-Precision model on all ten target groups. It flags fewer examples as hate, but when it does flag one, it is more often correct. Third, \texttt{Qwen3-4B} has low Recall, from \(0.24\) for Muslim targets to \(0.52\) for LGBTQ+ targets, which means it often defaults to the non-hate class. AUC also varies by target group, and the variation is systematic rather than model-specific. Averaged over the eight models, detection is strongest for LGBTQ+ (\(0.69\)), Christian (\(0.67\)) and Latinx (\(0.65\)) targets, and weakest for Women (\(0.57\)), Black (\(0.58\)), Muslim (\(0.59\)) and White (\(0.59\)) targets. Every one of the eight models has its best AUC on either LGBTQ+ or Christian targets and its worst on either Women or Black targets, so the ordering is not driven by any single model; the bottom four groups lie within \(0.02\) of each other, however, and their internal ranking should not be over-read. The gap that matters is the \(0.12\) between the best and worst group, which is larger than the \(0.08\) that separates the best and second-best Qwen model averaged over groups (\texttt{Qwen3-32B} \(0.68\) versus \texttt{Qwen3-8B} \(0.60\)). The global maximum is AUC \(0.77\) for \texttt{Qwen3-32B} on LGBTQ+ targets, and the global minimum, AUC \(0.48\) for \texttt{gemma-3-1b-it} on Black targets, is at chance level. This target dependence is consistent with prior work showing that hate-speech classifiers and annotations vary by target identity \citep{yoder-etal-2022-hate,sap-etal-2019-risk}.

\begin{table*}[h!]
\centering
\small
\setlength{\tabcolsep}{4pt}
\caption{Per-target hate-speech detection performance under the \textbf{base} (no-persona) condition. Models are columns; each target group contributes four metric rows. Each cell pools the 30 base prompt configurations that sampled that target group, i.e.\ 39{,}960 item-level predictions. Precision, Recall, and F1 are computed at a $0.5$ decision threshold; AUC is threshold-free. \textbf{Bold} marks the best score per row.}
\label{tab:per_target_metrics}
\begin{tabular}{l l cccccccc}
\toprule
Target & Metric & \multicolumn{4}{c}{Gemma~3} & \multicolumn{4}{c}{Qwen~3} \\
\cmidrule(lr){3-6} \cmidrule(lr){7-10}
 &  & 1B & 4B & 12B & 27B & 4B & 8B & 14B & 32B \\
\midrule
\multirow{4}{*}{Asian} & Precision & 0.50 & 0.57 & 0.62 & 0.63 & 0.63 & 0.56 & \textbf{0.66} & 0.61 \\
 & Recall & 0.71 & 0.51 & 0.65 & \textbf{0.73} & 0.29 & 0.60 & 0.48 & 0.71 \\
 & F1 & 0.59 & 0.54 & 0.63 & \textbf{0.68} & 0.39 & 0.58 & 0.55 & 0.66 \\
 & AUC & 0.51 & 0.59 & 0.67 & \textbf{0.70} & 0.60 & 0.59 & 0.68 & 0.69 \\\hline
\addlinespace[2pt]
\multirow{4}{*}{Black} & Precision & 0.49 & 0.52 & 0.57 & 0.56 & 0.58 & 0.53 & \textbf{0.61} & 0.55 \\
 & Recall & 0.56 & 0.60 & 0.71 & \textbf{0.80} & 0.30 & 0.66 & 0.52 & 0.78 \\
 & F1 & 0.52 & 0.56 & 0.63 & \textbf{0.66} & 0.40 & 0.59 & 0.56 & 0.65 \\
 & AUC & 0.48 & 0.54 & 0.63 & \textbf{0.64} & 0.54 & 0.56 & 0.63 & 0.62 \\\hline
\addlinespace[2pt]
\multirow{4}{*}{Christians} & Precision & 0.51 & 0.58 & 0.70 & 0.71 & 0.71 & 0.61 & \textbf{0.72} & 0.68 \\
 & Recall & \textbf{0.72} & 0.69 & 0.44 & 0.60 & 0.29 & 0.67 & 0.35 & 0.57 \\
 & F1 & 0.60 & 0.63 & 0.54 & \textbf{0.65} & 0.41 & 0.64 & 0.47 & 0.62 \\
 & AUC & 0.52 & 0.62 & 0.69 & \textbf{0.75} & 0.66 & 0.67 & 0.72 & 0.73 \\\hline
\addlinespace[2pt]
\multirow{4}{*}{Jews} & Precision & 0.50 & 0.56 & 0.62 & 0.61 & 0.62 & 0.55 & \textbf{0.70} & 0.61 \\
 & Recall & 0.69 & 0.65 & 0.73 & \textbf{0.85} & 0.32 & 0.73 & 0.52 & 0.74 \\
 & F1 & 0.58 & 0.60 & 0.67 & \textbf{0.71} & 0.42 & 0.63 & 0.60 & 0.67 \\
 & AUC & 0.50 & 0.60 & 0.69 & 0.71 & 0.59 & 0.60 & \textbf{0.71} & 0.70 \\\hline
\addlinespace[2pt]
\multirow{4}{*}{LGBTQ+} & Precision & 0.54 & 0.57 & 0.61 & 0.59 & 0.66 & 0.55 & \textbf{0.70} & 0.60 \\
 & Recall & 0.67 & 0.78 & 0.89 & \textbf{0.96} & 0.52 & 0.89 & 0.70 & 0.93 \\
 & F1 & 0.59 & 0.66 & 0.72 & \textbf{0.74} & 0.59 & 0.68 & 0.70 & 0.73 \\
 & AUC & 0.56 & 0.64 & 0.74 & 0.74 & 0.68 & 0.65 & 0.75 & \textbf{0.77} \\\hline
\addlinespace[2pt]
\multirow{4}{*}{Latinx} & Precision & 0.50 & 0.58 & 0.63 & 0.64 & 0.62 & 0.56 & \textbf{0.65} & 0.63 \\
 & Recall & 0.66 & 0.65 & 0.73 & \textbf{0.83} & 0.32 & 0.71 & 0.54 & 0.77 \\
 & F1 & 0.57 & 0.61 & 0.68 & \textbf{0.72} & 0.42 & 0.63 & 0.59 & 0.69 \\
 & AUC & 0.51 & 0.62 & 0.71 & \textbf{0.73} & 0.61 & 0.61 & 0.70 & 0.72 \\\hline
\addlinespace[2pt]
\multirow{4}{*}{Men} & Precision & 0.53 & 0.58 & 0.64 & 0.64 & 0.66 & 0.56 & \textbf{0.68} & 0.61 \\
 & Recall & 0.50 & 0.55 & 0.58 & 0.68 & 0.41 & \textbf{0.77} & 0.50 & 0.68 \\
 & F1 & 0.51 & 0.57 & 0.61 & \textbf{0.66} & 0.51 & 0.65 & 0.58 & 0.64 \\
 & AUC & 0.54 & 0.60 & 0.67 & \textbf{0.70} & 0.66 & 0.62 & 0.69 & 0.69 \\\hline
\addlinespace[2pt]
\multirow{4}{*}{Muslims} & Precision & 0.51 & 0.56 & 0.57 & 0.58 & 0.60 & 0.54 & \textbf{0.62} & 0.58 \\
 & Recall & 0.60 & 0.54 & 0.52 & \textbf{0.69} & 0.24 & 0.64 & 0.38 & 0.64 \\
 & F1 & 0.55 & 0.55 & 0.54 & \textbf{0.63} & 0.34 & 0.59 & 0.47 & 0.61 \\
 & AUC & 0.52 & 0.57 & 0.59 & \textbf{0.64} & 0.56 & 0.58 & 0.63 & 0.63 \\\hline
\addlinespace[2pt]
\multirow{4}{*}{White} & Precision & 0.53 & 0.56 & 0.57 & 0.57 & 0.59 & 0.53 & \textbf{0.62} & 0.57 \\
 & Recall & \textbf{0.73} & 0.56 & 0.55 & 0.69 & 0.34 & 0.71 & 0.44 & 0.66 \\
 & F1 & 0.61 & 0.56 & 0.56 & \textbf{0.63} & 0.43 & 0.61 & 0.51 & 0.61 \\
 & AUC & 0.55 & 0.57 & 0.59 & \textbf{0.64} & 0.57 & 0.57 & 0.62 & 0.62 \\\hline
\addlinespace[2pt]
\multirow{4}{*}{Women} & Precision & 0.51 & 0.54 & 0.56 & 0.56 & 0.58 & 0.53 & \textbf{0.59} & 0.54 \\
 & Recall & 0.61 & 0.55 & 0.51 & 0.59 & 0.31 & \textbf{0.66} & 0.42 & 0.64 \\
 & F1 & 0.56 & 0.54 & 0.53 & 0.57 & 0.40 & \textbf{0.59} & 0.49 & 0.59 \\
 & AUC & 0.51 & 0.55 & 0.58 & 0.60 & 0.55 & 0.56 & \textbf{0.60} & 0.59 \\
\bottomrule
\end{tabular}
\end{table*}

A formal correlation between Political Compass coordinates (Section~\ref{sec:baseline}) and per-target detection performance using Spearman's $\rho$ is left to a forthcoming revision; the AUC ordering above is the strongest preliminary indicator of group-specific selective sensitivity. 

\subsection{Downstream Sentiment: IBM Topic Sentiment}
\label{sec:ibm_sentiment_results}
We next evaluate whether persona-conditioned political framing changes a target-dependent sentiment task. The IBM sentiment experiment contains 30 unique test topics, each labelled as positive or negative sentiment toward its topic target. 

\begin{figure*}[h!]
\centering
\includegraphics[width=1.0\textwidth]{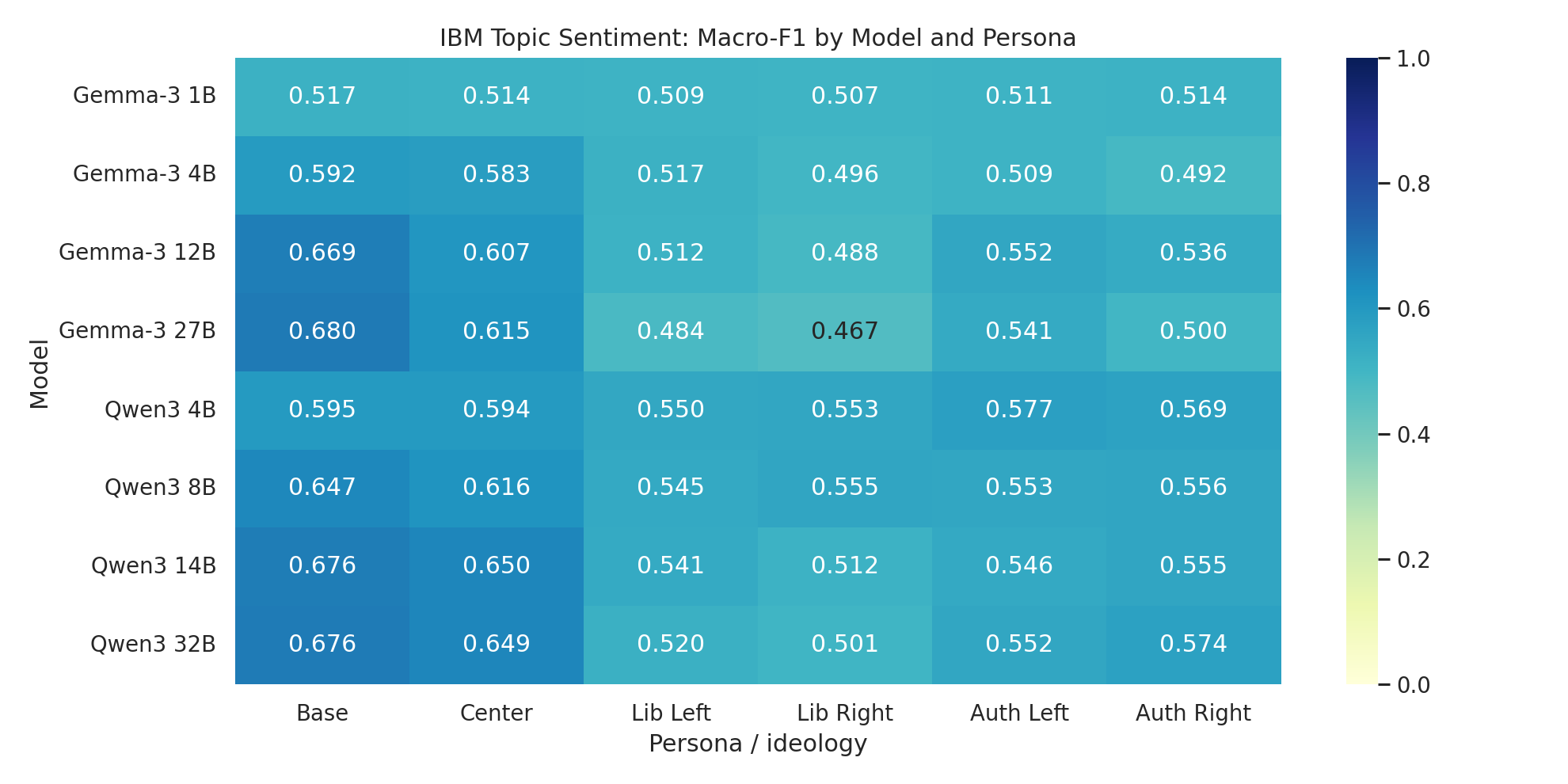}
\caption{IBM topic-level sentiment macro-F1 by model and persona condition. The base and centrist columns are consistently strongest, while explicit ideological personas usually reduce agreement with the topic-level gold sentiment labels. This should not be read as evidence that any ideological family is globally better at sentiment classification. It shows that persona conditioning changes task thresholds, and that stronger role prompts often hurt this small topic-level benchmark.}
\label{fig:ibm_sentiment_persona_heatmap}
\end{figure*}

\begin{figure*}[h!]
\centering
\includegraphics[width=1.0\textwidth]{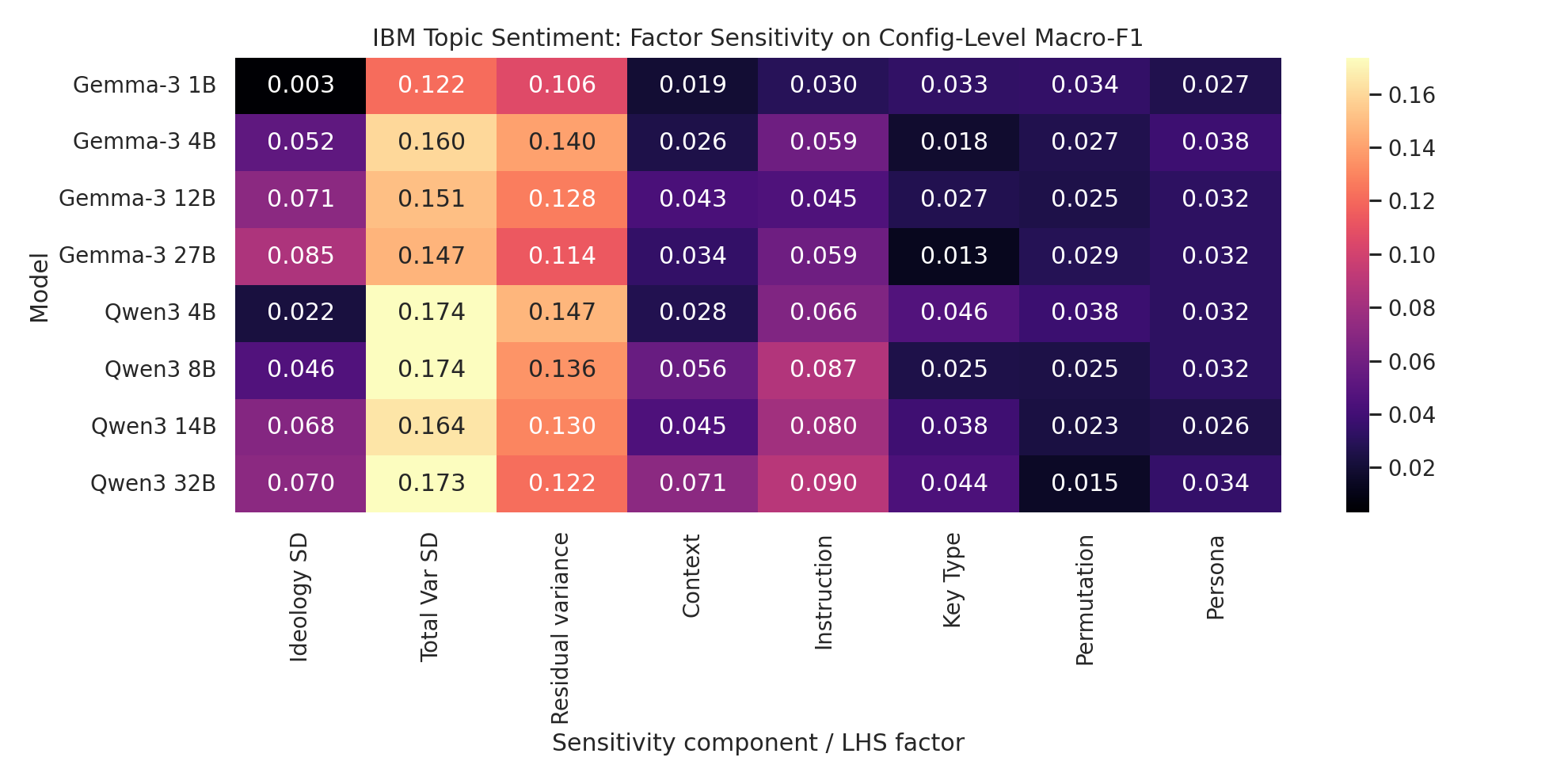}
\caption{IBM topic-level sentiment factor sensitivity for macro-F1. The analysis is computed from configuration-level macro-F1 across the base condition and all five persona conditions. `Ideology SD' captures the spread between persona-condition means, while the LHS factor columns are accumulated within persona strata. Cells report the standard-deviation-equivalent contribution of each component. The figure is used as a robustness diagnostic: it shows which prompt dimensions move downstream performance, rather than only whether a model is accurate in aggregate.}
\label{fig:ibm_sentiment_factor_sensitivity}
\end{figure*}

\begin{table*}[h!]
\centering
\small
\setlength{\tabcolsep}{4pt}
\caption{Overall and axis-level IBM topic-sentiment performance under the \textbf{base} (no-persona) condition. Rows group the 30 topic targets by the consensus taxonomy axis; support is shown as \((+\text{positive}/-\text{negative})\) topic counts. Precision, recall, and F1 are for the \texttt{POSITIVE} sentiment class, and AUC is computed from the model probability assigned to \texttt{POSITIVE}. AUC is undefined for Cross-cutting because all four Cross-cutting topics have positive gold sentiment labels; the precision value for that axis is therefore also not directly comparable to mixed-label axes. The best value in each metric row is shown in bold.}
\label{tab:ibm_sentiment_axis_metrics}
\begin{tabular}{l l cccccccc}
\toprule
Axis & Metric & \multicolumn{4}{c}{Gemma~3} & \multicolumn{4}{c}{Qwen~3} \\
\cmidrule(lr){3-6} \cmidrule(lr){7-10}
 & & 1B & 4B & 12B & 27B & 4B & 8B & 14B & 32B \\
\midrule
\multirow{4}{*}{All (+19/-11)} & Precision & 0.654 & 0.784 & 0.840 & \textbf{0.872} & 0.742 & 0.764 & 0.810 & 0.797 \\
 & Recall & 0.546 & 0.492 & 0.594 & 0.582 & 0.567 & 0.671 & 0.650 & \textbf{0.676} \\
 & F1 & 0.595 & 0.605 & 0.696 & 0.698 & 0.643 & 0.714 & 0.722 & \textbf{0.732} \\
 & AUC & 0.540 & 0.695 & 0.774 & \textbf{0.798} & 0.669 & 0.708 & 0.756 & 0.765 \\\hline
\addlinespace[2pt]
\multirow{4}{*}{Social (+8/-6)} & Precision & 0.606 & 0.743 & 0.839 & \textbf{0.859} & 0.722 & 0.719 & 0.801 & 0.786 \\
 & Recall & 0.550 & 0.492 & 0.630 & 0.600 & 0.545 & 0.667 & 0.677 & \textbf{0.706} \\
 & F1 & 0.577 & 0.592 & 0.720 & 0.706 & 0.621 & 0.692 & 0.734 & \textbf{0.744} \\
 & AUC & 0.550 & 0.706 & \textbf{0.815} & 0.811 & 0.699 & 0.720 & 0.793 & 0.805 \\\hline
\addlinespace[2pt]
\multirow{4}{*}{Political (+5/-2)} & Precision & 0.702 & 0.786 & 0.801 & \textbf{0.812} & 0.728 & 0.780 & 0.796 & 0.788 \\
 & Recall & 0.525 & 0.453 & 0.415 & 0.403 & 0.516 & \textbf{0.597} & 0.530 & 0.566 \\
 & F1 & 0.601 & 0.574 & 0.546 & 0.538 & 0.604 & \textbf{0.677} & 0.636 & 0.659 \\
 & AUC & 0.502 & 0.602 & 0.626 & \textbf{0.650} & 0.530 & 0.608 & 0.627 & 0.643 \\\hline
\addlinespace[2pt]
\multirow{4}{*}{Economic (+2/-3)} & Precision & 0.419 & 0.567 & 0.622 & \textbf{0.741} & 0.486 & 0.566 & 0.558 & 0.556 \\
 & Recall & 0.547 & 0.395 & 0.540 & 0.520 & 0.522 & \textbf{0.663} & 0.547 & 0.615 \\
 & F1 & 0.475 & 0.466 & 0.578 & \textbf{0.611} & 0.503 & \textbf{0.611} & 0.552 & 0.584 \\
 & AUC & 0.538 & 0.662 & 0.729 & \textbf{0.770} & 0.624 & 0.710 & 0.692 & 0.711 \\\hline
\addlinespace[2pt]
\multirow{4}{*}{Cross-cutting (+4/-0)} & Precision & \textbf{1.000} & \textbf{1.000} & \textbf{1.000} & \textbf{1.000} & \textbf{1.000} & \textbf{1.000} & \textbf{1.000} & \textbf{1.000} \\
 & Recall & 0.564 & 0.591 & 0.771 & \textbf{0.801} & 0.697 & 0.773 & 0.799 & 0.786 \\
 & F1 & 0.721 & 0.743 & 0.871 & \textbf{0.889} & 0.821 & 0.872 & 0.888 & 0.880 \\
 & AUC & NA & NA & NA & NA & NA & NA & NA & NA \\
\bottomrule
\end{tabular}
\end{table*}

In Figure~\ref{fig:ibm_sentiment_persona_heatmap}, we report aggregated model performance. Base prompting performs best on average, centrist prompting is usually close behind, and stronger ideological personas lower macro-F1 for nearly every model. This is not a contradiction with the Political Compass results. In the Political Compass task, moving the model with a persona prompt is the thing being measured. In IBM sentiment, the goal is to match a gold topic-sentiment label. A strong political role can pull the model toward its own issue priors, so agreement with the gold label drops.

Table~\ref{tab:ibm_sentiment_axis_metrics} disaggregates the base condition overall and by the target-taxonomy axis. In the overall row, Gemma 27B has the highest positive-class precision and AUC,
while Qwen 32B has the highest positive-class recall and F1. Social topics show the clearest scale-related gains, with AUC rising from 0.55 for Gemma 1B to above 0.80 for the largest Gemma and Qwen models. Political topics are more difficult: positive-class recall remains low for several larger models despite high precision, indicating conservative positive sentiment thresholds on these topics. The Cross-cutting row should be read separately because it contains only positive-gold topics.

Figure~\ref{fig:ibm_sentiment_factor_sensitivity} shows that downstream performance is itself prompt-sensitive. Residual variance remains large, but instruction, context, key type, permutation, and persona wording each move macro-F1 to different degrees across model families. This effect has also been demonstrated before in other domains~\citep{khattab2024dspy}. The between-condition spread, reported as \emph{Ideology SD}, increases with checkpoint size within both families. By contrast, the \emph{Persona} column measures wording differences within a persona condition and remains much smaller. Thus, larger checkpoints show greater performance separation between political roles, while the particular wording of a given role has a more limited effect.

The topic-level delta analysis clarifies what the heatmap does and does not imply. Averaged across models and topics, every collapsed ideological persona family performs below the base condition: Libertarian \(-0.117\), Right \(-0.111\), Left \(-0.099\), and Authoritarian \(-0.093\) in topic-level accuracy. Center is also slightly below base (\(-0.026\)), but it remains much closer to base than the stronger ideological roles. The small Authoritarian-minus-Libertarian contrast (\(+0.024\)) is therefore not evidence that authoritarian personas are good sentiment classifiers. It only says that, in this small benchmark, authoritarian personas are slightly less harmful than libertarian personas on average, and the uncertainty interval crosses zero.

The axis split helps explain where that average comes from. Political and economic topics push the Authoritarian-minus-Libertarian contrast upward, while social and cross-cutting topics dampen or reverse it. Put plainly, persona prompts do not make errors worse in one uniform direction. They move mistakes around. On these 30 topics, authoritarian personas lose less performance on some political and economic targets, while social topics are mixed and include clear cases in the opposite direction. Because the support counts are small, this is descriptive error analysis rather than a domain-level fairness claim.

The target-level diagnostic shows why aggregate sentiment scores need to be read carefully. Larger models perform strongly on many topics with clearer majority sentiment, such as multiculturalism, endangered species, trade-union bargaining, and intellectual property rights. They remain fragile on other topics where the gold label is less aligned with the model's apparent prior, such as mandatory retirement age, bribery, abstinence-only sex education, and voter photo ID. These examples illustrate that the downstream effect is not only a change in overall macro-F1, but also a redistribution of errors across politically meaningful topics.

\section{Discussion and Conclusions}
\label{sec:discussion}
We conducted a robust study on political behaviour in eight open-weight \LLMs across 14 languages, three quantization levels, and three different elicitation protocols. The main methodological result is simple: one pass through a political questionnaire is not enough. Prompt wording, language, answer format, and elicitation mode can all move the recovered coordinates. LHS sampling and factor decomposition do not reveal a hidden internal ideology. They give a more honest behavioural estimate: where the model lands under a specified evaluation design, and how much that estimate moves when the design changes.

The results show that larger models are usually recovered as left-libertarian in the base condition. The smallest models stay near the origin, but, as our item-level consistency analysis clearly shows, that should not be read as principled centrism, but rather as a functional ignorance, i.e., a failure to engage with the ideological content of the propositions. Larger models can be steered into distinct persona clusters, which suggests that they understand enough of the propositions and persona instructions for the task to recover a signal. Language is one of the largest external factors, especially on the economic axis; instruction phrasing and permutation also matter in some social-axis analyses. Quantization has smaller effects for the larger models and is overall not a significant factor. The Authoritarian-Left persona remains hardest to elicit and the chat diagnostic shows that writing an intermediate answer can change the coordinates.

\paragraph{The measurement problem:}
Our most fundamental finding is that a single pass through the \PCT is an unreliable estimator of a model's political position. The economic axis, derived from only 18 questions, has intrinsically higher per-test variance; layered on top of this, language, instruction phrasing, and answer-key format each contribute statistically significant additional variance for most models. Stable estimates require averaging over a broad sample of measurement configurations, which is achievable at reasonable computational cost. This has implications beyond political bias measurement: any psychometric evaluation of \LLMs that relies on a single-pass or low-sample protocol may be substantially less reliable than it appears, particularly under deterministic decoding, which can suppress distributional variance while amplifying prompt-sensitivity (Section~\ref{sec:uncertainty_table}).

\paragraph{Language is not a neutral carrier:}
Language choice is a major external factor and a clear argument against single-language political evaluation. Models can appear to hold meaningfully different recovered coordinates depending on the language of administration. We interpret this cautiously as a mixture of translation effects, uneven multilingual concept representation, and English-centric alignment transfer, rather than as direct evidence of culture-specific political adaptation.

\paragraph{Reasoning mode is not a simple safety dampener:}
The combined analysis does not support the claim that thinking uniformly moderates political stance. Chat thinking differs clearly from the aligned MCQ reference slice, and Qwen \texttt{think} mode shifts economic scores leftward relative to \texttt{no\_think}. The social-axis effect is smaller and varies across model sizes. The safer conclusion is that explicit reasoning changes the elicitation condition; it can move persona-conditioned coordinates, but it does not reliably suppress ideological expression.

\paragraph{Functional inaccessibility and the limits of current alignment:}
The relative difficulty of eliciting Authoritarian-Left position points to an asymmetry in current alignment behaviour, but the mechanism remains open. It may reflect under-representation in the training corpora, safety training, refusal heuristics, annotator demographics, or a mismatch between the PCT's historical propositions and the model's learned political abstractions. We therefore avoid treating this as direct evidence of over-refusal without accompanying text-level analysis. What the current results do show is behavioural: standard persona prompting is less able to move models into this region than into other regions of the compass.

\paragraph{Downstream consequences:}
The downstream experiments show why the measurement problem matters, but they should be framed as sensitivity analyses rather than as a complete fairness audit. In hate-speech detection, persona and ideology account for at most \(0.13\) SD in \(P(\text{Hate})\), against a target-group spread in AUC of \(0.57\) to \(0.69\) and a clear scaling effect with model size, suggesting that task behaviour is more anchored than the PCT persona clusters alone would imply. In IBM topic-level sentiment, persona prompting often reduces agreement with the gold topic sentiment, especially for non-centrist personas. The new topic-level diagnostic shows that this degradation is issue-specific: political and economic topics tend to favour authoritarian over libertarian personas, whereas social topics are mixed and include strong libertarian-over-authoritarian cases. These results do not prove a simple mapping from compass coordinate to downstream harm. They do show that political role prompting can change classification thresholds and performance in sensitive tasks, providing an empirical bridge between abstract political coordinates and practical model behaviour.

\section{Limitations and Future Work}
\label{sec:limitations_future}
\begin{itemize}
\item \textbf{Instrument scope.} The \PCT was designed primarily for a Western political context. Its two axes may not capture the ideological landscape equally well across all 14 tested languages, and the questionnaire has limited coverage of contemporary political issues.

\item \textbf{Scoring reconstruction.} The reconstructed \PCT scoring function is high fidelity, with held-out RMSE near 0.003 on the \([-10,+10]\) scale, but it remains an approximation to a proprietary
black-box scoring rule.

\item \textbf{Sampling and interactions.} LHS gives strong marginal coverage of the prompt space, but it does not exhaust the full factorial space and does not identify every high-order interaction. Future work
should add targeted follow-up designs around the largest interactions found here.

\item \textbf{Computational cost.} The method is deliberately repetitive: every sampled configuration must be evaluated across every item, model, persona, and, when applicable, quantization level. This makes the estimator more stable, but the cost scales linearly with the number of benchmark items and can become expensive for large downstream datasets. This is why IBM sentiment is treated as a 30-topic experiment, while claim-level stance and hate-speech settings require separate compute planning.

\item \textbf{MCQ/chat comparability and protocol isolation.} The MCQ/chat comparison is constrained by the small exact overlap between prompt spaces. The standard-chat contrasts should therefore be interpreted with model-cluster caveats. The denser Qwen comparison shows that the complete \texttt{think} and \texttt{no\_think} protocols produce different recovered distributions, but it does not isolate reasoning mode from their different temperature, top-$p$, and token-budget settings. Future work should use a fully crossed design with identical prompt rows and decoding settings, varying only whether the reasoning mode is enabled. Additional diversity and more repeated generation samples would also help separate decoding variability from prompt-induced variability.

\item \textbf{Chat explanation and answer recovery.} Our text audit shows that explicit refusal is rare and that the Stage~2 answer-mapping procedure usually agrees with an explicit final stance. However, the predefined qualifying-language cue list is not a validated measure of hedging, and automated closed LLM judges that we experimented with do not agree reliably on subjective properties such as persona fidelity, moral distancing, or rationale consistency. A dedicated blinded human annotation study should evaluate a representative sample of explanations with a preregistered codebook and reported inter-annotator agreement. Future work should also compare same-model answer mapping with human labels and independent classifiers, particularly for the two smallest Gemma checkpoints where Stage~1--Stage~2 agreement is lower.

\item \textbf{Language and cultural interpretation.} Language is varied in the present design, but country, nationality, and respondent identity are not. Future experiments should cross language with an explicitly stated country or speaker context, use independently validated translations, and compare multiple countries within the same language. This would help distinguish tokenization and translation effects from genuinely country-conditioned political behaviour.

\item \textbf{Downstream task scope.} IBM sentiment contains only 30 unique topic items, so axis and target-domain analyses are descriptive. The stance datasets and hate-speech target groups provide broader tests, but a stronger fairness claim would require matched counterfactual examples and explicit group-level causal designs.

\item \textbf{Model scope.} The study covers Gemma~3 and Qwen~3 model families. Generalisability to other architectures, alignment recipes, pretraining corpora, and deployment settings remains open.

\end{itemize}

\section*{Acknowledgments}
The authors acknowledge the computing resources provided by the L3i laboratory at La Rochelle Université, with support from the French government and the Nouvelle-Aquitaine Region.

\section*{Declarations}
\label{declarations}
\begin{itemize}
    \item \textbf{Funding:} This work was supported by the Slovenian Research and Innovation Agency through the core research programme Knowledge Technologies (No. P2-0103), the project Embeddings-based Techniques for Media Monitoring Applications (EMMA, No. L2-50070), and the project Large Language Models for Digital Humanities (LLM4DH, No. GC-0002). Further support was provided by the ACTUADA project (No. 2022-2021-17014610), funded by the Nouvelle-Aquitaine Region, France, and by the European Union through the HORIZON-WIDERA-2023-TALENTS-01-01 grant AI4DH (No. 101186647). The views and opinions expressed are those of the authors only and do not necessarily reflect those of the European Union. Neither the European Union nor the granting authority can be held responsible for them.
    \item \textbf{Competing Interests:} The authors declare that they have no competing interests.
    \item \textbf{Ethics Approval and Consent to Participate:} No new data from human participants were collected. This study evaluated computational models using existing research datasets under their applicable licenses. Ethics approval and participant consent were therefore not required.
    \item \textbf{Consent for Publication:} Not applicable.
    \item \textbf{Generative AI Use Disclosure:} Generative AI tools, specifically Gemini 2.5 Pro~\citep{comanici2025gemini25pushingfrontier}, Gemini 3.1 Pro~\citep{google2026gemini31pro}, Claude Sonnet 4 and Claude Opus 4~\citep{anthropic2025claude4}, Claude Sonnet 4.5~\citep{anthropic2025claudesonnet45}, Claude Opus 4.5~\citep{anthropic2025claudeopus45}, GPT-4o~\citep{openai2024gpt4ocard}, GPT-5.5~\citep{openai2026gpt55}, and Grok 3~\citep{xai2025grok3}, were used to assist with the development workflow, including boilerplate code generation, Bash scripting for experiment automation, and debugging. Generative AI tools were also used for limited language editing and refinement of manuscript text. All AI-generated or AI-assisted material was reviewed, verified, and, where necessary, revised by the authors. The authors remain fully responsible for the content, implementation, analysis, and conclusions presented in this work.
    \item \textbf{Code and Data Availability:} The code and scripts used to reproduce the experiments presented in this work are publicly available in the \href{https://github.com/nishan-chatterjee/llm-bias-detection}{project repository}. The datasets and associated resources are publicly available through the \href{https://huggingface.co/datasets/nishan-chatterjee/llm-bias-detection}{Hugging Face dataset repository}. Versioned snapshots of the code and data corresponding to the final version of the manuscript will be archived in a persistent repository, and the resulting persistent identifiers will be added to the article.
    \item \textbf{Author Contributions:} L.D.: Conceptualization, Methodology, Software, Validation, Formal Analysis, Investigation, Data Curation, Visualization, Writing--Original Draft, Writing--Review \& Editing. N.C.: Conceptualization, Methodology, Software, Validation, Formal Analysis, Investigation, Data Curation, Visualization, Writing--Original Draft, Writing--Review \& Editing, Project Administration. M.M.: Conceptualization, Methodology, Software, Validation, Formal Analysis, Investigation, Data Curation, Visualization, Writing--Original Draft, Writing--Review \& Editing. S.P.: Conceptualization, Supervision, Writing--Review \& Editing. A.D.: Conceptualization, Supervision, Writing--Review \& Editing. Luka Debevc and Nishan Chatterjee contributed equally to this work. All authors approved the final manuscript and agree to be accountable for all aspects of the work.
\end{itemize}

\bibliography{bibliography}

\begin{thebibliography}{}

\bibitem[{Anthropic}, 2025a]{anthropic2025claude4}
{Anthropic} (2025a).
\newblock {Claude 4} system card.
\newblock \url{https://www.anthropic.com/system-cards}.
\newblock Accessed: September 2, 2026.

\bibitem[{Anthropic}, 2025b]{anthropic2025claudeopus45}
{Anthropic} (2025b).
\newblock {Claude Opus 4.5} system card.
\newblock \url{https://www.anthropic.com/system-cards}.
\newblock Accessed: September 2, 2026.

\bibitem[{Anthropic}, 2025c]{anthropic2025claudesonnet45}
{Anthropic} (2025c).
\newblock {Claude Sonnet 4.5} system card.
\newblock \url{https://www.anthropic.com/system-cards}.
\newblock Accessed: September 2, 2026.

\bibitem[Argyle et~al., 2023]{argyle2023out}
Argyle, L.~P., Busby, E.~C., Fulda, N., Gubler, J.~R., Rytting, C., and
  Wingate, D. (2023).
\newblock Out of one, many: Using language models to simulate human samples.
\newblock {\em Political Analysis}, 31(3):337--351.

\bibitem[Badshah and Sajjad, 2024]{badshah2024quantifyingcapabilitiesllmsscale}
Badshah, S. and Sajjad, H. (2024).
\newblock Quantifying the capabilities of llms across scale and precision.

\bibitem[Bar-Haim et~al., 2017]{bar-haim-etal-2017-stance}
Bar-Haim, R., Bhattacharya, I., Dinuzzo, F., Saha, A., and Slonim, N. (2017).
\newblock Stance classification of context-dependent claims.
\newblock In {\em Proceedings of the 15th Conference of the European Chapter of
  the Association for Computational Linguistics: Volume 1, Long Papers}, pages
  251--261, Valencia, Spain. Association for Computational Linguistics.

\bibitem[Bolukbasi et~al., 2016]{bolukbasi2016man}
Bolukbasi, T., Chang, K.-W., Zou, J.~Y., Saligrama, V., and Kalai, A.~T.
  (2016).
\newblock Man is to computer programmer as woman is to homemaker? debiasing
  word embeddings.
\newblock {\em Advances in neural information processing systems}, 29.

\bibitem[Caliskan et~al., 2017]{caliskan2017semantics}
Caliskan, A., Bryson, J.~J., and Narayanan, A. (2017).
\newblock Semantics derived automatically from language corpora contain
  human-like biases.
\newblock {\em Science}, 356(6334):183--186.

\bibitem[Chao et~al., 2024]{chao2024jailbreakbenchopenrobustnessbenchmark}
Chao, P., Debenedetti, E., Robey, A., Andriushchenko, M., Croce, F., Sehwag,
  V., Dobriban, E., Flammarion, N., Pappas, G.~J., Tramer, F., Hassani, H., and
  Wong, E. (2024).
\newblock Jailbreakbench: An open robustness benchmark for jailbreaking large
  language models.

\bibitem[Davidson et~al., 2019]{davidson2019racial}
Davidson, T., Bhattacharya, D., and Weber, I. (2019).
\newblock Racial bias in hate speech and abusive language detection datasets.
\newblock {\em arXiv preprint arXiv:1905.12516}.

\bibitem[Dettmers et~al., 2022]{dettmers2022gpt3}
Dettmers, T., Lewis, M., Belkada, Y., and Zettlemoyer, L. (2022).
\newblock Gpt3. int8 (): 8-bit matrix multiplication for transformers at scale.
\newblock {\em Advances in neural information processing systems},
  35:30318--30332.

\bibitem[Durmus et~al., 2023]{durmus2023towards}
Durmus, E., Nguyen, K., Liao, T.~I., Schiefer, N., Askell, A., Bakhtin, A.,
  Chen, C., Hatfield-Dodds, Z., Hernandez, D., Joseph, N., et~al. (2023).
\newblock Towards measuring the representation of subjective global opinions in
  language models.
\newblock {\em arXiv preprint arXiv:2306.16388}.

\bibitem[Elbouanani et~al., 2025]{elbouanani2025analyzingpoliticalbiasllms}
Elbouanani, A., Dufraisse, E., and Popescu, A. (2025).
\newblock Analyzing political bias in llms via target-oriented sentiment
  classification.

\bibitem[Faulborn et~al., 2025]{faulborn-etal-2025-little}
Faulborn, M., Sen, I., Pellert, M., Spitz, A., and Garcia, D. (2025).
\newblock Only a little to the left: A theory-grounded measure of political
  bias in large language models.
\newblock In Che, W., Nabende, J., Shutova, E., and Pilehvar, M.~T., editors,
  {\em Proceedings of the 63rd Annual Meeting of the Association for
  Computational Linguistics (Volume 1: Long Papers)}, pages 31684--31704,
  Vienna, Austria. Association for Computational Linguistics.

\bibitem[Feng et~al., 2023]{feng-etal-2023-pretraining}
Feng, S., Park, C.~Y., Liu, Y., and Tsvetkov, Y. (2023).
\newblock From pretraining data to language models to downstream tasks:
  Tracking the trails of political biases leading to unfair {NLP} models.
\newblock In Rogers, A., Boyd-Graber, J., and Okazaki, N., editors, {\em
  Proceedings of the 61st Annual Meeting of the Association for Computational
  Linguistics (Volume 1: Long Papers)}, pages 11737--11762, Toronto, Canada.
  Association for Computational Linguistics.

\bibitem[Gajewska et~al.,
  2025]{gajewska2025algorithmicfairnessnlppersonainfused}
Gajewska, E., Derbent, A., Chudziak, J.~A., and Budzynska, K. (2025).
\newblock Algorithmic fairness in nlp: Persona-infused llms for human-centric
  hate speech detection.

\bibitem[Gerganov and llama.cpp contributors, 2023]{gerganov2023llamacpp}
Gerganov, G. and llama.cpp contributors (2023).
\newblock {llama.cpp}: Llm inference in c/c++.
\newblock \url{https://github.com/ggml-org/llama.cpp}.
\newblock Accessed: July 15, 2026.

\bibitem[{Google DeepMind}, 2025]{comanici2025gemini25pushingfrontier}
{Google DeepMind} (2025).
\newblock {Gemini 2.5}: Pushing the frontier with advanced reasoning,
  multimodality, long context, and next generation agentic capabilities.
\newblock {\em arXiv preprint}.

\bibitem[{Google DeepMind}, 2026]{google2026gemini31pro}
{Google DeepMind} (2026).
\newblock {Gemini 3.1 Pro} model card.
\newblock \url{https://deepmind.google/models/model-cards/gemini-3-1-pro/}.
\newblock Accessed: September 2, 2026.

\bibitem[Gurgurov et~al., 2025]{gurgurov2025multilingualpoliticalviewslarge}
Gurgurov, D., Trinley, K., Vykopal, I., van Genabith, J., Ostermann, S., and
  Zamparelli, R. (2025).
\newblock Multilingual political views of large language models: Identification
  and steering.

\bibitem[Hartmann et~al., 2023]{hartmann2023political}
Hartmann, J., Schwenzow, J., and Witte, M. (2023).
\newblock The political ideology of conversational ai: Converging evidence on
  chatgpt's pro-environmental, left-libertarian orientation.
\newblock {\em arXiv preprint arXiv:2301.01768}.

\bibitem[Hooker et~al., 2020]{hooker2020characterising}
Hooker, S., Moorosi, N., Clark, G., Bengio, S., and Denton, E. (2020).
\newblock Characterising bias in compressed models.
\newblock {\em arXiv preprint arXiv:2010.03058}.

\bibitem[Jiang et~al., 2024]{jiang2024peektokenbiaslarge}
Jiang, B., Xie, Y., Hao, Z., Wang, X., Mallick, T., Su, W.~J., Taylor, C.~J.,
  and Roth, D. (2024).
\newblock A peek into token bias: Large language models are not yet genuine
  reasoners.

\bibitem[Kamal et~al., 2025]{kamal2025detailed}
Kamal, S., Prakash, L. P.~Y., Rafiuddin, S., Rakib, M., Sen, A., and Choudhury,
  S.~R. (2025).
\newblock A detailed factor analysis for the political compass test: Navigating
  ideologies of large language models.
\newblock In {\em Proceedings of the 14th International Joint Conference on
  Natural Language Processing and the 4th Conference of the Asia-Pacific
  Chapter of the Association for Computational Linguistics}, pages 284--303.

\bibitem[Kamath et~al., 2025]{Kamath2025Gemma3T}
Kamath, G. T.~A. et~al. (2025).
\newblock Gemma 3 technical report.
\newblock {\em ArXiv}, abs/2503.19786.

\bibitem[Khattab et~al., 2024]{khattab2024dspy}
Khattab, O., Singhvi, A., Maheshwari, P., Zhang, Z., Santhanam, K.,
  Vardhamanan, S., Haq, S., Sharma, A., Joshi, T.~T., Moazam, H., Miller, H.,
  Zaharia, M., and Potts, C. (2024).
\newblock {DSPy}: Compiling declarative language model calls into
  self-improving pipelines.
\newblock In {\em The Twelfth International Conference on Learning
  Representations}.

\bibitem[Kim et~al., 2025]{kim2025linear}
Kim, J., Evans, J., and Schein, A. (2025).
\newblock Linear representations of political perspective emerge in large
  language models.
\newblock {\em arXiv preprint arXiv:2503.02080}.

\bibitem[Kumar et~al., 2024]{kumar2024scalinglawsprecision}
Kumar, T., Ankner, Z., Spector, B.~F., Bordelon, B., Muennighoff, N., Paul, M.,
  Pehlevan, C., Ré, C., and Raghunathan, A. (2024).
\newblock Scaling laws for precision.

\bibitem[Kwon et~al., 2023]{kwon2023efficient}
Kwon, W., Li, Z., Zhuang, S., Sheng, Y., Zheng, L., Yu, C.~H., Gonzalez, J.~E.,
  Zhang, H., and Stoica, I. (2023).
\newblock Efficient memory management for large language model serving with
  pagedattention.
\newblock In {\em Proceedings of the ACM SIGOPS 29th Symposium on Operating
  Systems Principles}.

\bibitem[Li et~al., 2026]{li2026analysingsafetypitfallssteering}
Li, Y., Fastowski, A., Zaradoukas, E., Prenkaj, B., and Kasneci, G. (2026).
\newblock Analysing the safety pitfalls of steering vectors.

\bibitem[Lutz et~al., 2025]{lutz2025promptmakespersonasystematic}
Lutz, M., Sen, I., Ahnert, G., Rogers, E., and Strohmaier, M. (2025).
\newblock The prompt makes the person(a): A systematic evaluation of
  sociodemographic persona prompting for large language models.

\bibitem[Luz~de Araujo et~al., 2026]{luz-de-araujo-etal-2026-persistent}
Luz~de Araujo, P.~H., Hedderich, M.~A., Modarressi, A., Schuetze, H., and Roth,
  B. (2026).
\newblock Persistent personas? role-playing, instruction following, and safety
  in extended interactions.
\newblock In Demberg, V., Inui, K., and Marquez, L., editors, {\em Proceedings
  of the 19th Conference of the {E}uropean Chapter of the {A}ssociation for
  {C}omputational {L}inguistics (Volume 1: Long Papers)}, pages 5329--5359,
  Rabat, Morocco. Association for Computational Linguistics.

\bibitem[Ma et~al., 2024]{ma2024visualroleplayuniversaljailbreakattack}
Ma, S., Luo, W., Wang, Y., and Liu, X. (2024).
\newblock Visual-roleplay: Universal jailbreak attack on multimodal large
  language models via role-playing image character.

\bibitem[McKay et~al., 1979]{McKay1979}
McKay, M.~D., Beckman, R.~J., and Conover, W.~J. (1979).
\newblock A comparison of three methods for selecting values of input variables
  in the analysis of output from a computer code.
\newblock {\em Technometrics}, 21(2):239.

\bibitem[Motoki et~al., 2024]{motoki2024more}
Motoki, F., Pinho~Neto, V., and Rodrigues, V. (2024).
\newblock More human than human: measuring chatgpt political bias.
\newblock {\em Public Choice}, 198(1):3--23.

\bibitem[{OpenAI}, 2024]{openai2024gpt4ocard}
{OpenAI} (2024).
\newblock {GPT-4o} system card.
\newblock {\em arXiv preprint}.

\bibitem[{OpenAI}, 2026]{openai2026gpt55}
{OpenAI} (2026).
\newblock {GPT-5.5} system card.
\newblock \url{https://openai.com/index/gpt-5-5-system-card/}.
\newblock Accessed: September 2, 2026.

\bibitem[{OpenRouter}, 2026]{openrouter}
{OpenRouter} (2026).
\newblock {OpenRouter}: A unified interface for llms.
\newblock \url{https://openrouter.ai}.
\newblock Accessed: July 15, 2026.

\bibitem[R{\"o}ttger et~al., 2024]{rottger2024political}
R{\"o}ttger, P., Hofmann, V., Pyatkin, V., Hinck, M., Kirk, H., Schuetze, H.,
  and Hovy, D. (2024).
\newblock Political compass or spinning arrow? towards more meaningful
  evaluations for values and opinions in large language models.
\newblock In {\em Proceedings of the 62nd Annual Meeting of the Association for
  Computational Linguistics (Volume 1: Long Papers)}, pages 15295--15311.

\bibitem[Rozado, 2024]{rozado2024political}
Rozado, D. (2024).
\newblock The political preferences of llms.
\newblock {\em PloS one}, 19(7):e0306621.

\bibitem[Rupprecht et~al.,
  2025]{rupprecht2025promptperturbationsrevealhumanlike}
Rupprecht, J., Ahnert, G., and Strohmaier, M. (2025).
\newblock Prompt perturbations reveal human-like biases in large language model
  survey responses.

\bibitem[Santurkar et~al., 2023]{santurkar2023whose}
Santurkar, S., Durmus, E., Ladhak, F., Lee, C., Liang, P., and Hashimoto, T.
  (2023).
\newblock Whose opinions do language models reflect?
\newblock In {\em International Conference on Machine Learning}, pages
  29971--30004. PMLR.

\bibitem[Sap et~al., 2019]{sap-etal-2019-risk}
Sap, M., Card, D., Gabriel, S., Choi, Y., and Smith, N.~A. (2019).
\newblock The risk of racial bias in hate speech detection.
\newblock In {\em Proceedings of the 57th Annual Meeting of the Association for
  Computational Linguistics}, pages 1668--1678, Florence, Italy. Association
  for Computational Linguistics.

\bibitem[Sap et~al., 2020]{sap2020social}
Sap, M., Gabriel, S., Qin, L., Jurafsky, D., Smith, N.~A., and Choi, Y. (2020).
\newblock Social bias frames: Reasoning about social and power implications of
  language.
\newblock In {\em Proceedings of the 58th annual meeting of the association for
  computational linguistics}, pages 5477--5490.

\bibitem[Sun et~al., 2024]{sun2024multiturncontextjailbreakattack}
Sun, X., Zhang, D., Yang, D., Zou, Q., and Li, H. (2024).
\newblock Multi-turn context jailbreak attack on large language models from
  first principles.

\bibitem[Templeton et~al., 2024]{templeton2024scaling}
Templeton, A., Conerly, T., Marcus, J., Lindsey, J., Bricken, T., Chen, B.,
  Pearce, A., Citro, C., Ameisen, E., Jones, A., Cunningham, H., Turner, N.~L.,
  McDougall, C., MacDiarmid, M., Freeman, C.~D., Sumers, T.~R., Rees, E.,
  Batson, J., Jermyn, A., Carter, S., Olah, C., and Henighan, T. (2024).
\newblock Scaling monosemanticity: Extracting interpretable features from
  claude 3 sonnet.
\newblock {\em Transformer Circuits Thread}.

\bibitem[Turpin et~al., 2023]{turpin2023language}
Turpin, M., Michael, J., Perez, E., and Bowman, S. (2023).
\newblock Language models don't always say what they think: Unfaithful
  explanations in chain-of-thought prompting.
\newblock {\em Advances in Neural Information Processing Systems},
  36:74952--74965.

\bibitem[{xAI}, 2025]{xai2025grok3}
{xAI} (2025).
\newblock {Grok 3 Beta}: The age of reasoning agents.
\newblock \url{https://x.ai/news/grok-3}.
\newblock Accessed: September 2, 2026.

\bibitem[Yang et~al., 2025]{yang2025qwen3technicalreport}
Yang, A. et~al. (2025).
\newblock Qwen3 technical report.

\bibitem[Yi et~al., 2024]{yi2024jailbreakattacksdefenseslarge}
Yi, S., Liu, Y., Sun, Z., Cong, T., He, X., Song, J., Xu, K., and Li, Q.
  (2024).
\newblock Jailbreak attacks and defenses against large language models: A
  survey.

\bibitem[Yoder et~al., 2022]{yoder-etal-2022-hate}
Yoder, M.~M., Ng, L. H.~X., Brown, D.~W., and Carley, K.~M. (2022).
\newblock How hate speech varies by target identity: A computational analysis.
\newblock In Fokkens, A. and Srikumar, V., editors, {\em Proceedings of the
  26th Conference on Computational Natural Language Learning (CoNLL)}, pages
  27--39, Abu Dhabi, United Arab Emirates (Hybrid). Association for
  Computational Linguistics.

\bibitem[Yuan et~al., 2025]{yuan2025hatefulpersonhatefulmodel}
Yuan, S., Nie, E., Tawfelis, M., Schmid, H., Sch{\"u}tze, H., and F{\"a}rber,
  M. (2025).
\newblock Hateful person or hateful model? investigating the role of personas
  in hate speech detection by large language models.

\bibitem[Zhang et~al., 2026]{zhang2026rethinkingpersonalizationlargelanguage}
Zhang, C., Lu, Y., Fang, L., Zheng, C., Chai, J., Wang, X., Yin, G., Lin, W.,
  Wang, Y., and Lin, Z. (2026).
\newblock Rethinking personalization in large language models at the token
  level.

\bibitem[Zhao et~al., 2025]{zhao2025llmsencodeharmfulnessrefusal}
Zhao, J., Huang, J., Wu, Z., Bau, D., and Shi, W. (2025).
\newblock Llms encode harmfulness and refusal separately.

\bibitem[Zheng et~al., 2024]{zheng2024largelanguagemodelsrobust}
Zheng, C., Zhou, H., Meng, F., Zhou, J., and Huang, M. (2024).
\newblock Large language models are not robust multiple choice selectors.

\bibitem[Zou et~al., 2025]{zou2025representationengineeringtopdownapproach}
Zou, A., Phan, L., Chen, S., Campbell, J., Guo, P., Ren, R., Pan, A., Yin, X.,
  Mazeika, M., Dombrowski, A.-K., Goel, S., Li, N., Byun, M.~J., Wang, Z.,
  Mallen, A., Basart, S., Koyejo, S., Song, D., Fredrikson, M., Kolter, J.~Z.,
  and Hendrycks, D. (2025).
\newblock Representation engineering: A top-down approach to ai transparency.

\end{thebibliography}
\onecolumn
\appendix
\section{English MCQ/Chat Robustness Analysis}
\label{app:joint_mcq_chat}

\begin{figure}[p]
\centering
\includegraphics[width=0.88\textwidth]{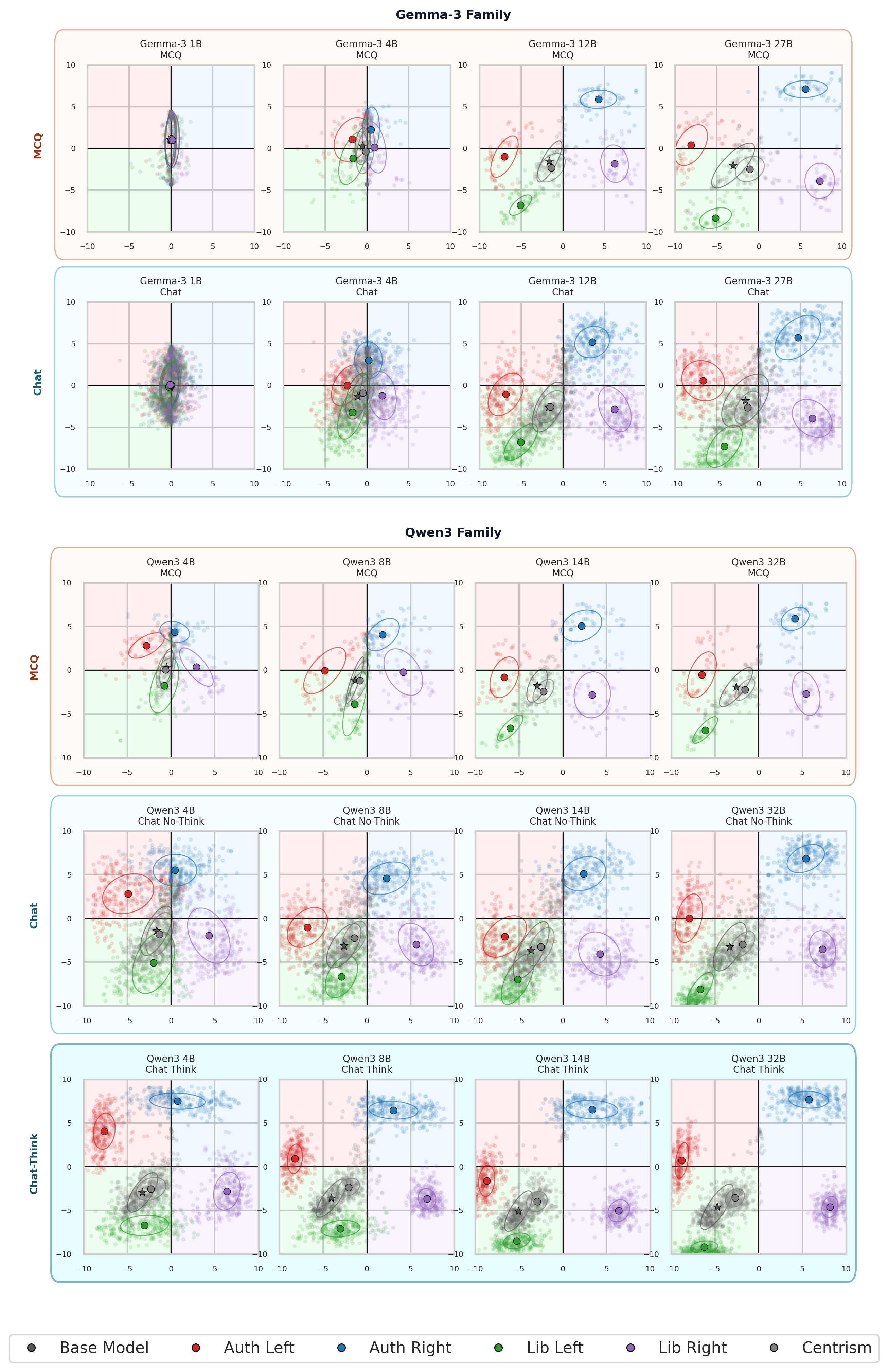}
\caption{English-only Political Compass coordinates under direct scoring and chat. Each row is one asking-and-measurement procedure and each column is one model size. Each point is a sampled configuration; the ellipse summarizes the spread of points for one persona. The figure is a comparison of direct scoring/MCQ vs the chat-then-score and a visual diagnostic of whether asking for written text before classification changes the recovered clusters. The English MCQ reference slice and chat prompt spaces are closely related but not fully identical, so it should not be treated as a stand-alone causal comparison.}
\label{fig:mcq_chat_positions}
\end{figure}

Here we use the English-only MCQ/chat comparison because it is the cleanest visual diagnostic of protocol differences (see Figure~\ref{fig:mcq_chat_positions}). The two prompt spaces are still related rather than identical, so this comparison should not be interpreted as a fully crossed causal protocol experiment. Exact matching across the two designs yields only 12 MCQ-linked pairs per base model, and MCQ-vs-chat contrasts are therefore treated as measurement-protocol diagnostics.

\begin{figure}[p]
\centering
\includegraphics[width=1.0\textwidth]{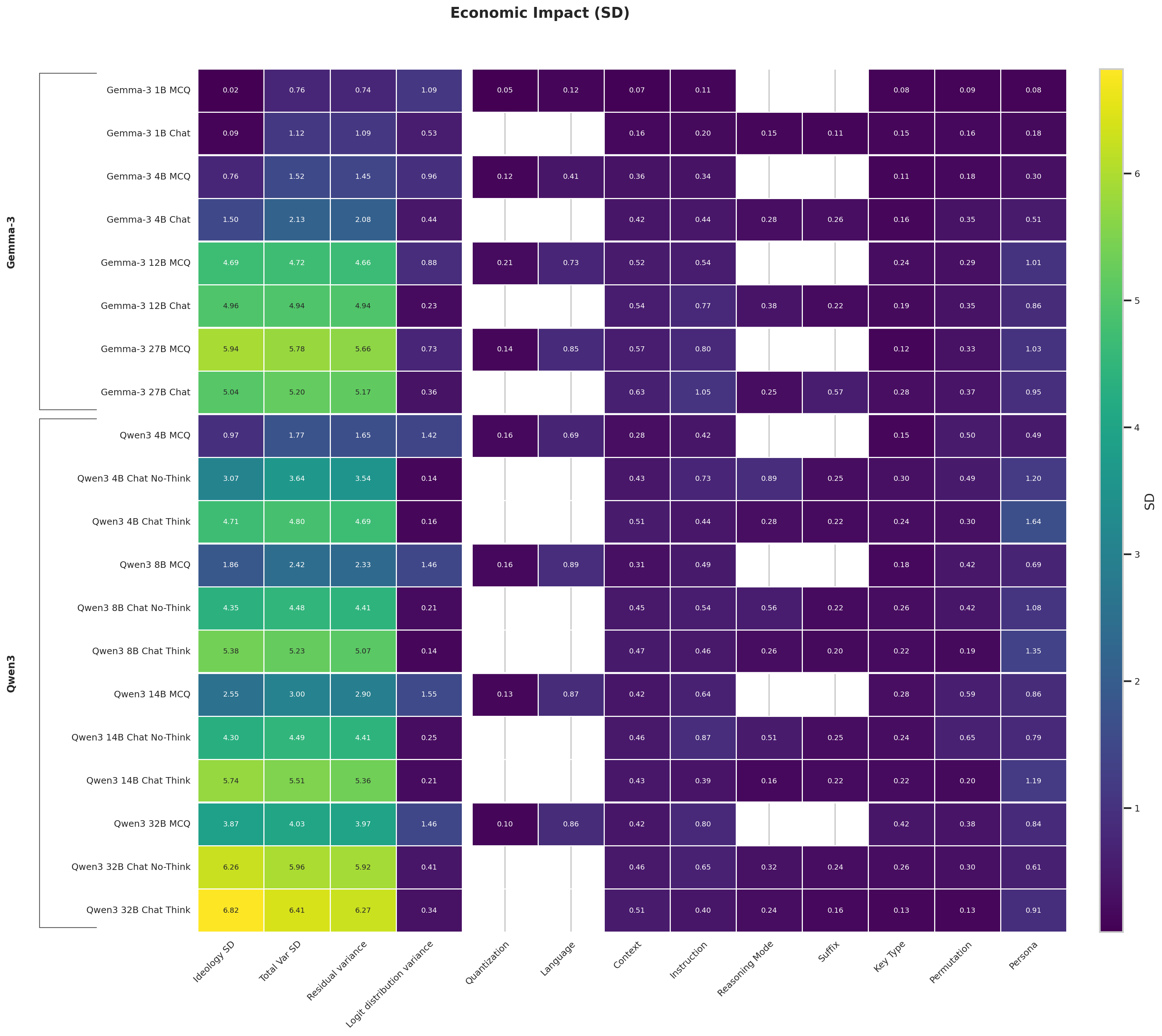}
\caption{Economic-axis factor-sensitivity diagnostic for direct scoring and chat. Each row combines one model with one asking-and-measurement procedure. A cell gives the estimated standard-deviation-equivalent change in the recovered economic coordinate associated with the named factor while larger values indicate more variation across that factor's settings. Blank cells mark the factors that do not apply to that procedure. Because the MCQ and chat prompt spaces are closely related but not identical, this figure can be used to interpret the comparisons as a robustness diagnostic rather than a formal factorial significance test.}
\label{fig:combined_sensitivity_economic}
\end{figure}

The strongest aligned contrast is the chat-thinking comparison against the MCQ reference slice. Across matched pairs, Chat Think shifts economic coordinates leftward by $-1.91$ compass points and social coordinates libertarian by $-1.54$ points relative to MCQ, with both contrasts surviving permutation testing and Benjamini-Hochberg correction ($q<0.001$). Standard chat also differs from MCQ at the pair level (economic: $-0.65$, $q=0.008$; social: $-0.72$, $q=0.043$), but the model-cluster intervals for these standard-chat contrasts reach zero. Thus the standard-chat result is suggestive, while the chat-thinking result is the stronger protocol effect. The relative factor sensitivities underlying these shifts on the economic axis are detailed in Figure~\ref{fig:combined_sensitivity_economic}.

\begin{figure}[p]
\centering
\includegraphics[width=1.0\textwidth]{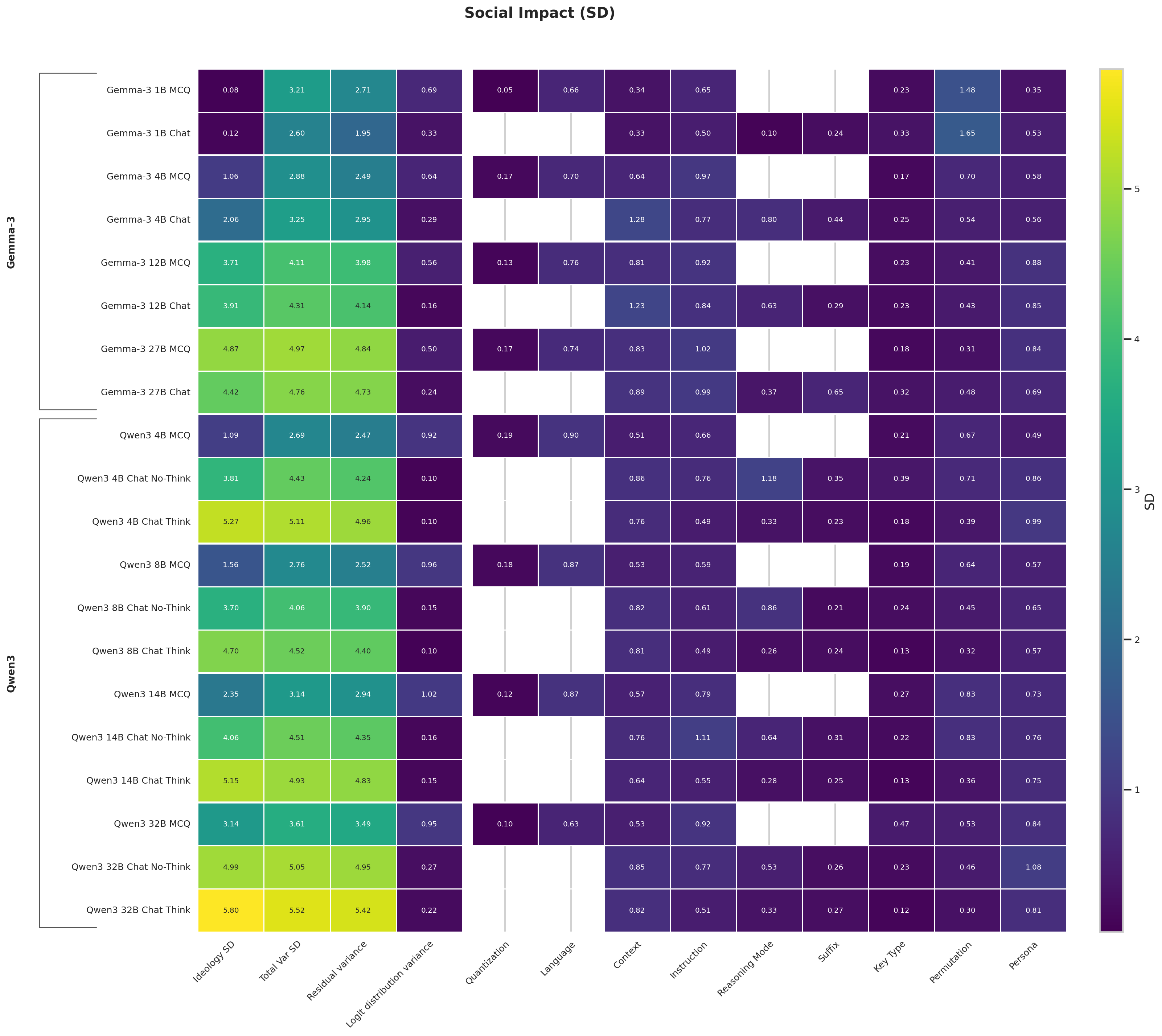}
\caption{Social-axis factor-sensitivity diagnostic for direct scoring and chat. Values are calculated in the same way as in Figure~\ref{fig:combined_sensitivity_economic}. The figure shows how much the recovered social coordinate changes across the applicable settings for each model and procedure.}
\label{fig:combined_sensitivity_social}
\end{figure}

Within Qwen, where \texttt{think} and \texttt{no\_think} runs can be matched much more densely, thinking shifts economic scores leftward by $-0.33$ compass points on average ($q<0.001$; model-cluster CI $[-0.56,-0.18]$). This is the clearest matched protocol effect in the appendix. The corresponding social-axis difference is smaller and varies by model: the average is $-0.20$ points, but the model-cluster interval includes zero and Qwen3-8B moves in the opposite, more authoritarian direction (see Figure~\ref{fig:combined_sensitivity_social}). Explicit reasoning therefore changes the measurement, especially on the economic axis, but does not move every coordinate in one common direction.

\section{Mathematical Formulation of Coordinate Scoring}
\label{app:math}
The \PCT scoring function $f$ is linear: it maps a vector of responses to a scalar coordinate via a fixed weight vector. A key property of linear functions is that expectation commutes with the scoring function, so the score of the expected response equals the average of the individual scores.  This allows aggregation across configurations in any order without bias.

Let $V_q \mid s$ denote the random variable encoding the model's answer to question $q$ under setup $s$, and let $S$ be the set of sampled configurations.  The expected axis coordinate is:
\begin{align}
C_{\text{axis}}
&= f\bigl(\mathbb{E}[V]\bigr)
\tag{Expected score}\\
&\approx f \left( \frac{1}{|S|} \sum_{s \in S} \mathbb{E}[V \mid s] \right)
\tag{Sample mean approximation}\\
&= \frac{1}{|S|} \sum_{s \in S} \mathbb{E} \bigl[ f(V) \mid s \bigr]
\tag{Commutativity}\\
&= \frac{1}{|S|} \sum_{s \in S} \mathbb{E} \biggl[ \, \sum_{q \in Q} f_q(V_q) \biggm| s \biggr]
\tag{Decomposition by question}\\
&= \frac{1}{|S|} \sum_{s \in S} \sum_{q \in Q} f_q \bigl( \mathbb{E}[V_q \mid s] \bigr)
\tag{Linearity of $\mathbb{E}$ and $f_q$}\\
&= \frac{1}{|S|} \sum_{s \in S} \sum_{q \in Q} f_q(\mathbf{p}_{q,s})
\tag{Substitution of probabilities}
\end{align}
where $f_q$ is the linear weight for question q and $\mathbf{p}_{q,s}$ is the exact conditional  distribution from the forward pass. Because $f_q$ is linear and $\mathbf{p}_{q,s}$ is the exact distribution, this estimator is unbiased with respect to the sampled configurations.

\section{Hyperparameters Used}
\label{sec:appendix-hyperparameters}
Table~\ref{tab:hyperparameters} reports the settings needed to reproduce the reported runs. Direct scoring has no text-decoding temperature because it reads the candidate answer probabilities from one forward pass. The chat diagnostic is the only experiment that samples generated text and therefore needs sampling.

\begin{table}[!htbp]
\centering
\small
\setlength{\tabcolsep}{4pt}
\renewcommand{\arraystretch}{1.15}
\caption{Run settings for the reported experiments. Sampling parameters used are presented as $[T,p,k,L]$, where $T$ is temperature, $p$ is top-$p$, $k$ is top-$k$, and $L$ is the maximum number of new tokens. N/A indicates that the experiment uses direct scoring and does not sample any generated text. These are fixed evaluation settings based on inference guidelines set by model providers and what users would most likely use while using cloud-based APIs like ~\cite{openrouter} or local solutions using llama.cpp~\citep{gerganov2023llamacpp} and VLLM~\citep{kwon2023efficient}, and not parameters tuned on an evaluation set.}
\label{tab:hyperparameters}
\begin{tabularx}{\textwidth}{@{} >{\raggedright\arraybackslash}p{0.25\textwidth} p{0.36\textwidth} X @{}}
\toprule
\textbf{Experiment} & \textbf{Sampling settings} & \textbf{Additional settings} \\
\midrule
Political Compass,\newline direct scoring & N/A & 300 Latin-hypercube configurations; direct-scoring protocol (Figure~\ref{fig:mcq_pipeline}). \\
\addlinespace
Political Compass,\newline chat diagnostic & \begin{tabular}[t]{@{}l@{\hspace{4pt}}l@{}}
Gemma-3:     & $[1.0,\,0.95,\,64,\,4096]$ \\
Qwen3 no\_think: & $[0.7,\,0.80,\,20,\,4096]$ \\
Qwen3 think: & $[0.6,\,0.95,\,20,\,8192]$
\end{tabular} & 300 Latin-hypercube configurations; chat protocol (Figure~\ref{fig:chat_pipeline}). Generation batch size 32; candidate-scoring batch size 128. \\
\addlinespace
IBM topic sentiment & N/A & 300 Latin-hypercube configurations; direct-scoring protocol (Figure~\ref{fig:mcq_pipeline}). \\
\addlinespace
Hate speech & N/A & 300 Latin-hypercube configurations; direct-scoring protocol (Figure~\ref{fig:mcq_pipeline}). \\
\bottomrule
\end{tabularx}
\end{table}

\section{Computational Budget}
\label{sec:appendix-budget}
The computational footprint spans both direct-scoring evaluation and open-ended generation runs across four core experiments:

\begin{itemize}
    \item \textbf{Direct scoring (Political Compass):} $\approx$40 GPU-hours across 40\,GB A100 GPUs ($\approx$1,600 GB-hours). Compute remains low because candidate-answer probabilities are extracted from single forward passes without generation.
    \item \textbf{Chat diagnostic:} $\approx$672 total GPU-hours ($\approx$41,472 GB-hours), comprising:
    \begin{itemize}
        \item 4$\times$ A40 GPUs (48\,GB) for 4 days ($\approx$384 GPU-hours; 18,432 GB-hours)
        \item 3$\times$ H100 GPUs (80\,GB) for 4 days ($\approx$288 GPU-hours; 23,040 GB-hours)
    \end{itemize}
    \item \textbf{IBM topic sentiment:} 3$\times$ H100 GPUs (80\,GB) for 4 days ($\approx$288 GPU-hours; 23,040 GB-hours).
    \item \textbf{Hate speech:} 3$\times$ H100 GPUs (80\,GB) for 4 days ($\approx$288 GPU-hours; 23,040 GB-hours).
\end{itemize}

In total, the logged runs accounted for approximately \textbf{1,288 GPU-hours} (roughly \textbf{89,150 VRAM GB-hours}) across 14 dedicated GPUs. This total excludes exploratory pilots, hardware-failure reruns, and offline notebook analysis.

\clearpage

\clearpage

\clearpage

\end{document}